\documentclass{WileyMSP-template}

\usepackage{textcomp}
\usepackage{graphicx}
\usepackage{float}
\usepackage[english]{babel}
\usepackage{amssymb}
\usepackage[normalem]{ulem}
\usepackage{amsmath}
\usepackage{amsthm}
\usepackage{amsfonts}
\usepackage{enumerate}
\usepackage{mathtools}
\usepackage{hyperref} 
\usepackage[dvipsnames]{xcolor}

\usepackage{lineno}
\usepackage{bm}

\usepackage{algorithmic}
\usepackage{algorithm}
\usepackage{ragged2e}
\usepackage{cite}

\begin{document}

\pagestyle{fancy}
\rhead{\includegraphics[width=2.5cm]{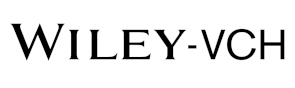}}

\title{From Folding Mechanics to Robotic Function: A Unified Modeling Framework for Compliant Origami}

\maketitle


\author{Bohan Zhang,}
\author{Bo Wang$^*$,}
\author{Huajiang Ouyang,}
\author{Zhigang Wu,}
\author{Haohao Bi,}
\author{Jiawei Xu,}
\\
\author{Mingchao Liu$^*$,}
\author{Weicheng Huang$^*$}

\begin{affiliations}

$*$ Corresponding authors. \\

\hfill \break

Bohan Zhang\\
Department of Engineering Mechanics,Northwestern Polytechnical University, \\ Xi’an, 710072, China \\

\hfill \break
Bo Wang\\
Department of Engineering Mechanics,Northwestern Polytechnical University,  Xi’an, 710072, China\\
Shenzhen Research Institute of Northwestern Polytechnical University, Shenzhen, 518063, China \\
bowang@nwpu.edu.cn (Corresponding Author) 

\hfill \break

Huajiang Ouyang\\
School of Mechanical Engineering, Southwest Jiaotong University, Chengdu, 610031, China \\
School of Engineering, University of Liverpool, Liverpool, L69 3GH, U.K.

\hfill \break

Zhigang Wu\\
School of Aeronautics and Astronautics, Sun Yat-sen University, No.66 Gongchang Road, Guangming District, Shenzhen, 518107, Guangdong, China

\hfill \break

Haohao Bi\\
School of Science, Qingdao University of Technology, Qingdao, 266520, China

\hfill \break

Jiawei Xu\\
Department of Engineering Mechanics, Northwestern Polytechnical University, \\ Xi’an, 710072, China \\

\hfill \break

Mingchao Liu\\
Department of Mechanical Engineering, University of Birmingham, Birmingham, B15 2TT, UK\\
m.liu.2@bham.ac.uk (Corresponding Author)

\hfill \break

Weicheng Huang\\
School of Engineering, Newcastle University, Newcastle upon Tyne, NE1 7RU, UK\\
weicheng.huang@newcastle.ac.uk (Corresponding Author)

\hfill \break

\end{affiliations}



\begin{abstract}

\begin{justify}

\textbf{Abstract:} Origami-inspired architectures offer a powerful route toward lightweight, reconfigurable, and programmable robotic systems.
Yet, a unified mechanics framework capable of seamlessly bridging rigid folding, elastic deformation, and stability-driven transitions in compliant origami remains lacking.
Here, we introduce a geometry-consistent modeling framework based on discrete differential geometry (DDG) that unifies panel elasticity and crease rotation within a single variational formulation.
By embedding crease–panel coupling directly into a mid-edge geometric discretization, the framework naturally captures rigid-folding limits, distributed bending, multistability, and nonlinear dynamic snap-through within one mechanically consistent structure.
This unified description enables programmable control of stability and deformation across rigid and compliant regimes, allowing origami structures to transition from static folding mechanisms to active robotic modules.
An implicit dynamic formulation incorporating gravity, contact, friction, and magnetic actuation further supports strongly coupled multiphysics simulations.
Through representative examples spanning single-fold bifurcation, deployable Miura membranes, bistable Waterbomb modules, and Kresling-based crawling robots, we demonstrate how geometry-driven mechanics directly informs robotic functionality.
This work establishes discrete differential geometry as a foundational design language for intelligent origami robotics, enabling predictive modeling, stability programming, and mechanics-guided robotic actuation within a unified computational platform.

\end{justify}

\end{abstract}

\begin{justify}


\section{Introduction}

Origami-inspired architectures have emerged as a powerful paradigm for designing lightweight, compact, and highly reconfigurable robotic systems \cite{rus2018design,meloni2021engineering}.
By encoding large shape transformations into geometric folding patterns, origami structures enable deployable space systems \cite{yue2023review}, adaptive mechanical metamaterials \cite{huang2024integration,li2025demand}, robotic actuators \cite{jin2021origami}, and bio-inspired locomotion platforms \cite{fang2017origami}. 
In recent years, the integration of origami principles with soft robotics has enabled robots capable of rapid deployment, multistable actuation, and programmable morphing, offering new opportunities for intelligent machines that exploit structural mechanics as a functional design element \cite{ze2022soft,yang2023morphing,yan2023origami}. 
Despite these advances, predicting and controlling the mechanics of compliant origami systems remains challenging due to the strong coupling between folding kinematics, panel elasticity, and nonlinear stability transitions \cite{misseroni2024origami}.

Existing modeling strategies for origami structures can be broadly categorized into several paradigms according to their underlying mechanical assumptions \cite{zhu2022review}.
First, rigid-panel models represent origami as assemblies of perfectly rigid facets connected by rotational hinges.
These approaches efficiently capture folding compatibility, deployment paths, and degrees of freedom in rigid origami \cite{tachi2009simulation,gattas2013miurabase}.
However, they usually neglect distributed bending and in-plane deformation of panels, limiting their applicability to compliant origami and soft robotic systems where elasticity plays an essential role.
Second, bar-and-hinge or bar-and-spring models approximate origami surfaces using skeletal networks \cite{liu2017nonlinear,vasudevan2024homogenization}.
In these formulations, bar elements represent in-plane stretching, while auxiliary springs or diagonal elements mimic bending resistance, and crease lines are assigned rotational springs to capture folding behavior.
Although these models partially incorporate structural flexibility, their mechanical response often relies on heuristic stiffness assignments and lacks a direct geometric description of the underlying surface.
Third, continuum-based approaches, most commonly finite element methods (FEM), provide a general framework for modeling thin structures undergoing large deformation \cite{wang2023simulation,wang2024deployment,li2026homogenization}.
While such formulations capture both stretching and bending of origami panels, folding creases require specialized treatments to accommodate large rotations and geometric discontinuities, and the resulting simulations are often computationally intensive.
Consequently, bridging rigid folding kinematics with distributed shell deformation within a unified and geometrically consistent framework remains an outstanding challenge.

A key difficulty in modeling compliant origami lies in the intrinsic geometric nature of folding mechanics.
Folding transforms a flat sheet through large rotations while maintaining compatibility between panels and creases, generating complex energy landscapes that govern bistability, multistability, and snap-through transitions \cite{waitukaitis2015origami,mora2025programming,hu2026peculiar}.
Capturing these behaviors requires formulations that simultaneously account for geometric constraints, curvature-driven shell mechanics, and crease-induced rotations. 
This challenge becomes particularly significant as origami structures evolve from deployable mechanisms toward functional robotic devices, which require models capable of capturing compliant deformation, distributed elasticity, and nonlinear dynamics within a unified and computationally efficient framework \cite{rus2018design,hawkes2010programmable,zhang2021programmable}.
However, existing modeling paradigms rarely satisfy these requirements while preserving both geometric structure and mechanical consistency.

Discrete differential geometry (DDG), which directly discretizes geometric invariants such as metric and curvature, offers a natural foundation for bridging geometry and mechanics in thin structures \cite{huang2025tutorial,tong2026discrete}.
By preserving intrinsic geometric relationships, DDG-based formulations enable robust handling of large rotations and deformations while maintaining a consistent connection between surface geometry and mechanical response \cite{korner2021simple}. 
In this work, we develop a unified DDG-based modeling framework for compliant origami structures with direct relevance to origami robotics.
Panels and creases are represented within a common mid-edge geometric discretization that encodes both surface curvature and fold rotations.
Panel mechanics are governed by discrete curvature measures derived from shell theory, while crease mechanics are described through dihedral-angle variables capturing localized rotations.
This unified variational formulation provides a geometry-consistent description of stretching, bending, and crease rotation within a single mechanical framework.

The proposed framework further supports dynamic simulation of active origami systems through an implicit time-integration scheme incorporating multiphysics interactions, including gravity, contact, friction, and magnetic actuation.
This capability enables direct modeling of the strongly coupled mechanics underlying origami robotic functionality, from bistable actuation modules to multistable locomotion mechanisms.
Through representative examples spanning fundamental folding units, deployable Miura sheets \cite{schenk2013geometry}, bistable Waterbomb structures \cite{ma2020folding}, and Kresling-based crawling robots \cite{ze2022soft}, we demonstrate how the DDG framework bridges rigid and compliant origami regimes and enables predictive modeling of stability-driven robotic behaviors.
By establishing a geometry-consistent computational foundation for origami mechanics, this work provides a principled framework for the analysis, design, and control of next-generation origami robotic systems.

\section{A Geometry-Consistent Discrete Mechanics Framework for Compliant Origami }
\label{sec:simulation}

This section presents a geometry-consistent discrete mechanics framework for compliant origami. 
Rather than treating folding kinematics and elastic deformation separately, the formulation is constructed within a unified variational setting that remains valid from rigid-folding limits to fully compliant behavior. 
The mechanical response of the system is governed by the total potential energy, which consistently accounts for panel deformation and crease mechanics within the same geometric description. 
Denoting the generalized coordinates by $\mathbf{q}$, the total elastic potential energy is defined as
\begin{equation}
\Pi(\mathbf{q}) = U_{\text{shell}}(\mathbf{q}) + U_{\text{crease}}(\mathbf{q}),
\label{eq:potential}
\end{equation}
where $U_{\text{shell}}$ represents the stretching and bending energies of the deformable panels, discretized using a discrete differential geometry (DDG) formulation \cite{huang2025tutorial}, while $U_{\text{crease}}$ accounts for rotational stiffness and localized bending along the fold lines.
This energetic decomposition enables both rigid-folding and compliant-deformation behaviors to be incorporated within a single mechanical framework.
To capture transient behaviors, we adopt a dynamic formulation that consistently couples elastic and inertial effects.
Accordingly, the system dynamics are derived from the Lagrangian
\begin{equation}
\mathcal{L}(\mathbf{q}, \dot{\mathbf{q}}) =
T(\dot{\mathbf{q}}) - \Pi(\mathbf{q}),
\label{eq:Lagrangian}
\end{equation}
where $\mathbf{q}$ and $\dot{\mathbf{q}}$ denote the generalized coordinates and velocities, respectively, $T$ is the kinetic energy, and $\Pi$ is the total potential energy defined above.
This unified variational formulation naturally captures rigid-folding motions, compliant deformations, and their dynamic coupling, thereby enabling the simulation of large-amplitude dynamics, stability transitions, and strongly nonlinear responses.

\subsection{Unified Geometric and Variational Formulation}


For realistic modeling of compliant origami robots, non-conservative external effects — including gravity, contact, friction, and field-driven actuation (e.g., magnetic forces) — must also be incorporated. By taking the variation of Equation~\eqref{eq:Lagrangian} with respect to the generalized coordinates and 
further incorporating non-conservative effects such as damping and external loading, 
the dynamic response of the origami structure is governed by

\begin{equation}
\mathbb{M} \, \ddot{\mathbf{q}}(t) 
+ \mathbb{C} \, \dot{\mathbf{q}}(t) 
+ \mathbb{K} \, \mathbf{q}(t) 
= \mathbf{f}_\text{ext}(t),
\label{eq:dynamic_full}
\end{equation}
where  
$\mathbb{M}$ is the mass matrix derived from the kinetic energy, 
$\mathbb{C}$ is the damping matrix accounting for energy dissipation, 
and $\mathbf{f}_\text{ext}(t)$ represents the externally applied forces. 
In Equation~\eqref{eq:dynamic_full}, the generalized coordinate vector $\mathbf{q}(t)$ collects all discrete degrees of freedom (DOFs) of the origami system. As illustrated in Figure~\ref{fig:element}, a flexible origami structure consists of two fundamental modules: the shell module and the crease module. A minimal origami unit contains at least one crease set and two adjacent shell panels. 
The detailed discretization procedures of these two modules will be presented in the following sections. Here, we directly outline the structure of the generalized coordinates. The DOFs of the origami system consist of two distinct sets of variables.
The first set corresponds to the translational degrees of freedom of the shell mesh, namely the spatial coordinates of all vertices,
\[
\mathbf{v}_{(a)}(t) =
\begin{bmatrix}
x_{(a)}(t)\quad y_{(a)}(t)\quad z_{(a)}(t)
\end{bmatrix},
\;\text{with} \; a = 1, \ldots, N_{1},
\]
where $N_{1}$   denotes the total number of vertices. 
The second set comprises the virtual folding angles associated with each crease element,
\[
\varphi_{v,(b)}(t), \;\text{with} \; b = 1, \ldots, N_{2},
\]
where $N_{2}$ is the total number of creases. 
The generalized coordinate vector is therefore written compactly as
\begin{equation}
\mathbf{q}(t) =
\left[
\mathbf{v}_{(1)}(t),
\dots,
\mathbf{v}_{(N_{1})}(t),
\varphi_{v,(1)}(t),
\dots,
\varphi_{v,(N_{2})}(t)
\right]^{\mathsf T},
\end{equation}
which yields a total of $3N_{1} + N_{2}$ degrees of freedom.
The configuration-dependent stiffness matrix $\mathbb{K}(\mathbf{q})$ is obtained from the second derivative of the elastic potential energy $\Pi(\mathbf{q})$ with respect to the generalized coordinates, namely
\[
\mathbb{K}(\mathbf{q}) = \frac{\partial^2 \Pi(\mathbf{q})}{\partial \mathbf{q}^2}.
\]
Owing to the geometric and material nonlinearities inherent in origami structures, 
$\mathbb{K}$ depends explicitly on the instantaneous configuration $\mathbf{q}(t)$ and is therefore continuously updated during deformation. 
Equation~\eqref{eq:dynamic_full} thus provides a unified formulation that accounts simultaneously for inertial, elastic, damping, and external effects within a unified nonlinear dynamic framework.
Time integration is performed using an implicit Euler scheme, leading to a nonlinear system solved by Newton iterations.
The source code of the simulation framework is publicly available in an open-access repository \cite{Zhang2026_compliant_origami_robot}.
%
%
In the following, we detail the construction of the elastic energy terms and the treatment of environmental interactions.

\subsection{Discrete Elastic Energy of Compliant Panels}

The deformable panels of the origami robot are modeled as thin shells, as illustrated in Figure~\ref{fig:element}(a).
Within the simulation framework, each crease is treated as a free boundary, and the two panels adjoining a crease are regarded as independent shell bodies.
Mechanical coupling between adjacent panels is introduced through dedicated crease elements.
As shown in Figure~\ref{fig:element}(b), each shell panel is represented as a smooth surface discretized by a triangular mesh.
To ensure a consistent interplay between geometry and mechanics, we adopt a discrete differential geometry (DDG) formulation \cite{huang2025tutorial,tong2026discrete}.
Under this framework, the surface is described by discrete analogues of the first and second fundamental forms, which characterize in-plane metric distortion and out-of-plane curvature, respectively.
These geometric descriptors provide the basis for defining membrane and bending strain measures in a manner that remains geometrically exact under large rotations and deformations.

\begin{figure}[t]
    \centering
    \includegraphics[
        width=0.9\linewidth,
        trim={0 10cm 0 0},
        clip
    ]{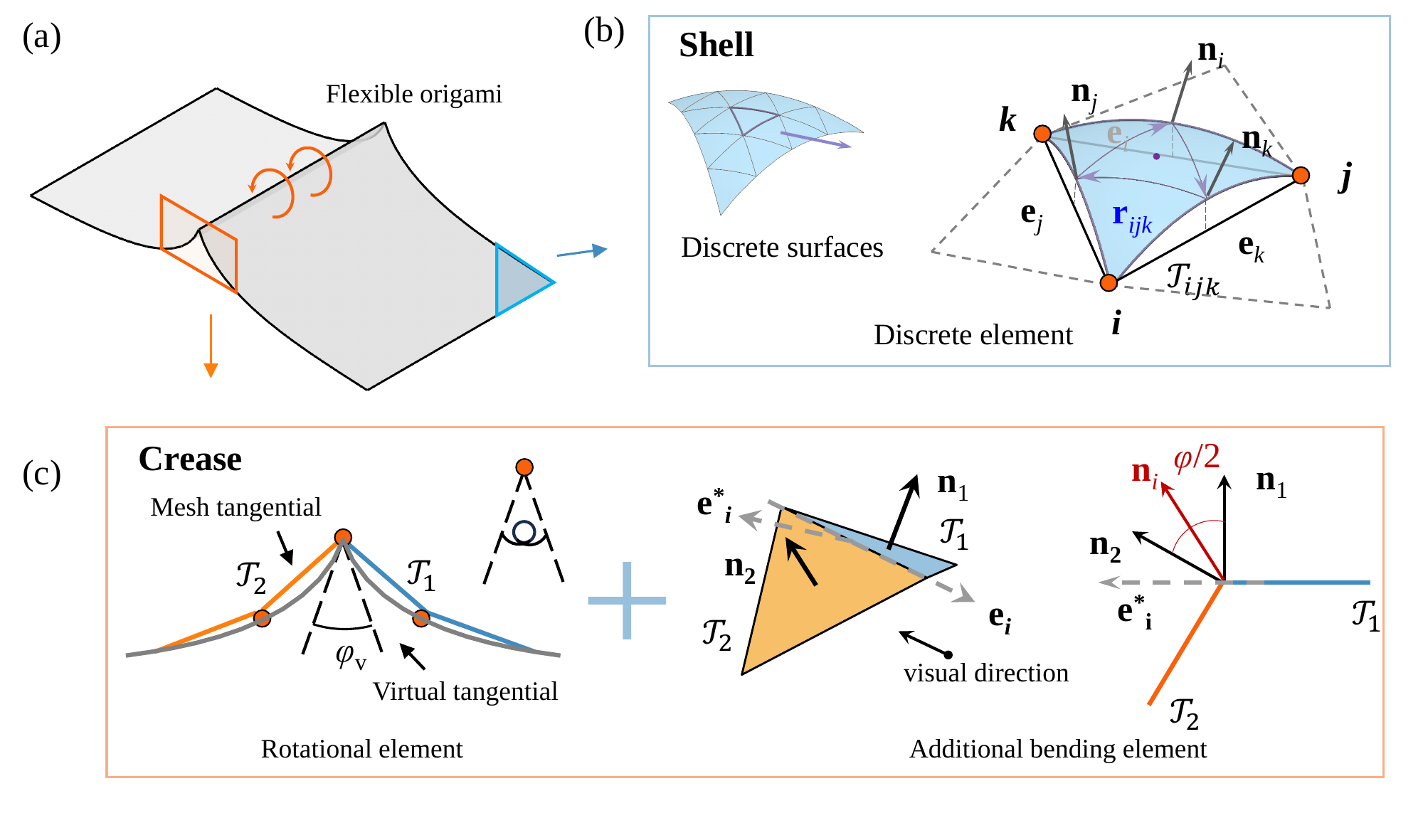}
    \caption {Schematic illustration of a flexible origami structure and its discrete modeling framework.
(a) Minimal structural composition of a flexible origami unit, consisting of a single crease (highlighted in orange) connecting two shell panels (highlighted in blue).
(b) Discretization of the shell module: the smooth panel surface is approximated by a triangular mesh, within which the discrete first and second fundamental forms are defined to characterize in-plane stretching and bending.
(c) Discretization of the crease module: the crease is modeled as a combination of rotational elements and additional bending elements.
 }
    \label{fig:element}
\end{figure}

To describe the deformation of each shell panel, consider a triangular element $\mathcal{T}_{ijk}$ with vertices $i$, $j$, and $k$, as shown in Figure~\ref{fig:element}(b). 
The shell panel is discretized into $N_{t}$ such triangular elements.
Both the discrete mesh and the associated smooth surface are embedded in $\mathbb{R}^3$. 
We use $\mathbf{e}_i$ denote the edge opposite to vertex $i$, 
and similarly for $\mathbf{e}_j$ and $\mathbf{e}_k$. 
If the discrete surface closely approximates the underlying smooth surface, the edge vectors provide finite-difference approximations of the tangent vectors of $\mathbf{r}_{ijk}$.
The first fundamental form measures the metric of the surface, i.e., the inner product of tangent vectors. 
In the discrete setting, this metric is naturally defined through edge vectors.
For an edge $\mathbf{e}_i$ opposite to vertex $i$, with vertex positions 
$\{ \mathbf{v}_i, \mathbf{v}_j, \mathbf{v}_k \}$, the discrete first fundamental form is defined as \cite{clarisse2012discrete}
\begin{equation}
\mathbb{A}(\mathbf{e}_i, \mathbf{e}_i) 
= (\mathbf{v}_k - \mathbf{v}_j)\cdot
(\mathbf{v}_k - \mathbf{v}_j ).
\end{equation}
Moreover, using the two edges emanating from vertex $i$,
$\mathbf{e}_k$ and $-\mathbf{e}_j$, as a basis spanning the discrete tangent space of the triangular element  $\mathcal{T}_{ijk}$, the discrete first fundamental form can be written in tensor form as
\begin{equation}\label{first form}
\mathbb{A}_{ijk} =
\begin{bmatrix}
(\mathbf{v}_j - \mathbf{v}_i) \cdot (\mathbf{v}_j - \mathbf{v}_i) 
&
(\mathbf{v}_j - \mathbf{v}_i) \cdot (\mathbf{v}_k - \mathbf{v}_i) \\[1mm]
(\mathbf{v}_k - \mathbf{v}_i) \cdot (\mathbf{v}_j - \mathbf{v}_i) 
&
(\mathbf{v}_k - \mathbf{v}_i) \cdot (\mathbf{v}_k - \mathbf{v}_i)
\end{bmatrix}.
\end{equation}
The second fundamental form characterizes the variation of the surface normal $\mathrm{d}\mathbf{n}$ induced by tangential displacement and measures the bending of the surface. 
To construct its discrete counterpart, a discrete normal field $\mathbf{n}$ needs be defined. 
We adopt a \emph{mid-edge discretization} in which the normal vector is associated with each edge midpoint\cite{grinspun2006discrete,chen2018physical}.
For an edge $\mathbf{e}_i$ of the triangular element $\mathcal{T}_{ijk}$, the associated mid-edge normal $\mathbf{n}_i$ is defined as follows: if $\mathbf{e}_i$ lies on the boundary of the structure, $\mathbf{n}_i$ coincides with the unit normal of the incident face,
\begin{equation}
\mathbf{n}_i =
\frac{(\mathbf{v}_j - \mathbf{v}_i)\times(\mathbf{v}_k - \mathbf{v}_i)}
{\left\|(\mathbf{v}_j - \mathbf{v}_i)\times(\mathbf{v}_k - \mathbf{v}_i)\right\|},
\end{equation}
otherwise, for an interior edge, $\mathbf{n}_i$ is taken as the average of the unit normals of the two adjacent faces sharing $\mathbf{e}_i$.
Within each triangular element, bending is evaluated through finite differences of the normal field along edge directions. 
Exploiting the fact that the segment connecting two edge midpoints is parallel to the opposite edge, the discrete 
second fundamental form along edge $\mathbf{e}_i$ is defined as
\begin{equation}
\mathbb{B}(\mathbf{e}_i, \mathbf{e}_i) 
:= 2 (\mathbf{n}_j - \mathbf{n}_k)\cdot 
(\mathbf{v}_k - \mathbf{v}_j ) .
\label{eq:primal-edge sff}
\end{equation}
And the tensor representation of the discrete second fundamental 
form for $\mathcal{T}_{ijk}$ is written as
\begin{equation}
\mathbb{B}_{ijk} =
2
\begin{bmatrix}
(\mathbf{n}_{j} - \mathbf{n}_{i}) \cdot (\mathbf{v}_{j} - \mathbf{v}_{i}) 
&
(\mathbf{n}_{j} - \mathbf{n}_{i}) \cdot (\mathbf{v}_{k} - \mathbf{v}_{i}) \\[6pt]
(\mathbf{n}_{k} - \mathbf{n}_{i}) \cdot (\mathbf{v}_{j} - \mathbf{v}_{i}) 
&
(\mathbf{n}_{k} - \mathbf{n}_{i}) \cdot (\mathbf{v}_{k} - \mathbf{v}_{i})
\end{bmatrix}.
\end{equation}
In this manner, the discrete fundamental forms provide geometric measures of in-plane metric distortion and out-of-plane curvature variation for each triangular element.
Given a reference configuration of the triangular mesh with vertex positions $\{ \bar{\mathbf{v}}_i, \bar{\mathbf{v}}_j, \bar{\mathbf{v}}_k \}$,  the corresponding reference first and second fundamental forms, $\bar{\mathbb{A}}_{ijk}$ and $\bar{\mathbb{B}}_{ijk}$, are constructed analogously.
Within the Kirchhoff–Love shell assumption, the deformation of a triangular element $\mathcal{T}_{ijk}$ is characterized by two strain measures: the membrane (in-plane) strain and the bending (out-of-plane) strain. 
The discrete membrane strain is defined as the relative change of the metric,
\begin{equation}
\boldsymbol{\varepsilon}_{ijk}
=
\bar{\mathbb{A}}_{ijk}^{-1}\mathbb{A}_{ijk}
-
\mathbf{I},
\end{equation}
which serves as a discrete analogue of the Green–Lagrange strain tensor\cite{armon2011geometry}.
The discrete bending strain is measured through the change of curvature,
\begin{equation}
\boldsymbol{\kappa}_{ijk}
=
\bar{\mathbb{A}}_{ijk}^{-1}
\left(
\mathbb{B}_{ijk}
-
\bar{\mathbb{B}}_{ijk}
\right).
\label{eq:bending strain}
\end{equation}
The total elastic energy of the shell is then written as the sum of stretching and bending contributions,
\begin{equation}
U_{\text{shell}}
=
U_{\text{stretch}}
+
U_{\text{bend}},
\end{equation}
with
\begin{equation}
U_{\text{stretch}}
=
\sum_{ijk}^{N_t}
\frac{h}{8}
\,
\left\| \boldsymbol{\varepsilon}_{ijk} \right\|_{SV}^2
\,
\sqrt{\det(\bar{\mathbb{A}}_{ijk})},
\end{equation}
\begin{equation}
U_{\text{bend}}
=
\sum_{ijk}^{N_t}
\frac{h^3}{24}
\,
\left\| \boldsymbol{\kappa}_{ijk} \right\|_{SV}^2
\,
\sqrt{\det(\bar{\mathbb{A}}_{ijk})}.
\label{eq:bending energy}
\end{equation}
Here, $h$ denotes the thickness of the shell and 
$\sqrt{\det(\bar{\mathbb{A}}_{ijk})}$ represents the reference area of the element. 
The operator $\left\| \cdot \right\|_{SV}^2$ 
indicates that the constitutive behavior follows the linear 
St.~Venant–Kirchhoff model. For instance, the membrane strain $\boldsymbol{\varepsilon}_{ijk}$ 
is evaluated as
\begin{equation}
\left\| \boldsymbol{\varepsilon}_{ijk} \right\|_{SV}^2
=
\frac{E \nu}{2(1 - \nu^2)}
\mathrm{Tr}^2(\boldsymbol{\varepsilon}_{ijk})
+
\frac{E}{2(1+\nu)}
\mathrm{Tr}(\boldsymbol{\varepsilon}_{ijk}^2),
\end{equation}
where $E$ and $\nu$ denote the Young's modulus and Poisson's ratio, respectively.

\subsection{Geometry-Consistent Modeling of Crease Interfaces}

We next consider the discrete formulation of crease in a flexible origami system.
Two independent shell panels are coupled along a prescribed crease line through a dedicated crease element.
The crease element is designed to consistently bridge the kinematics of smooth surfaces and their discrete triangular representations.
To this end, we decompose the crease contribution into two components: (i) a \emph{rotational element}, which governs the relative rotation between adjacent panels, and (ii) an \emph{additional bending element}, which accounts for the discrepancy between the discrete mesh geometry and the underlying smooth surface. 

We first introduce the formulation of the rotational element.
Figure~\ref{fig:element}(c) illustrates the cross-sectional profile of the compliant origami shell in the vicinity of the crease. 
Consider two adjacent triangular facets, denoted by $\mathcal{T}_1$ and $\mathcal{T}_2$, that share a common vertex located on the discrete crease line. 
In the discrete representation, geometric quantities are defined over triangular facets and, more specifically, along their edges. 
Within each facet $\mathcal{T}_1$ and $\mathcal{T}_2$, we connect the two vertices adjacent to the crease vertex, and the resulting line segment is referred to as the \emph{mesh tangential} direction. 
This direction characterizes the discrete geometric orientation of each panel near the crease.
To establish a consistent measure of the fold angle that reflects the underlying smooth geometry, we introduce a \emph{virtual tangential} direction on each side of the crease. 
As illustrated in Figure~\ref{fig:element}(c), the virtual tangentials are constructed from the smooth cross-sectional profile of the origami shell. 
Specifically, they are defined as the tangent directions of the underlying smooth surface at the crease location. 
Let $\mathbf{t}_1^{v}$ and $\mathbf{t}_2^{v}$ denote the virtual tangential directions associated with $\mathcal{T}_1$ and $\mathcal{T}_2$, respectively. 
The virtual fold angle $\varphi_v$ is defined as the signed dihedral angle between the corresponding orthogonal normal vectors, i.e.,
\begin{equation}
\varphi_v = \angle \big( \mathbf{n}_1^{v}, \mathbf{n}_2^{v} \big),
\label{eq:virtual fold angle}
\end{equation}
where $\mathbf{n}_i^{v}$ are unit vectors orthogonal to 
$\mathbf{t}_i^{v}$.
For clarity, tangential directions are illustrated 
in Figure~\ref{fig:element}(c), while the actual computation is performed using their orthogonal normals.
The rotational element is modeled as a torsional spring distributed along the crease. The elastic energy density per unit length is given by 
$ K_{\mathrm{r}} \left( \varphi_v - \varphi_0 \right)^2/2$, 
where $K_{\mathrm{r}}$ denotes the rotational stiffness and $\varphi_0$ is the rest angle. 
For a discrete crease segment shared by $\mathcal{T}_1$ and $\mathcal{T}_2$, the corresponding energy contribution is
\begin{equation}
U_{\text{rotation}}
=
\frac{1}{2} K_{\mathrm{r}} \left( \varphi_v - \varphi_0 \right)^2 \, \|\mathbf{e}_i\|,
\label{eq:rotation energy}
\end{equation}
where $ \|\mathbf{e}_i\|$ is the length of the common edge 
$\mathbf{e}_i$ between the two facets.
For a discretized crease composed of multiple edge segments, the total rotational energy is obtained by summing the contributions over all segments.
The additional bending element is introduced to characterize the bending of the facet pair adjacent to the crease. Since the rotational contribution is formulated in terms of the dihedral angle $\varphi_v$, it is natural to adopt a mid-edge discretization of the second fundamental form expressed 
in a dihedral-angle-based representation \cite{grinspun2006discrete}.
Similar to the definition of the virtual folding angle in Equation~\eqref{eq:virtual fold angle}, let $\mathbf{t}_1^{m}$ and $\mathbf{t}_2^{m}$ denote the mesh tangential directions extracted from the triangular facets. 
The  dihedral angle of  mesh tangential directions is defined through the orthogonal facet normals,
\begin{equation}
\varphi = \angle \big( \mathbf{n}_1^{m}, \mathbf{n}_2^{m} \big).
\end{equation}
In the dihedral-angle-based representation, the second fundamental form is expressed in terms of the dual edge direction $\mathbf{e}_i^{*}$, rather than the primal edge $\mathbf{e}_i$ adopted in Equation~\eqref{eq:primal-edge sff}.
The dual edge is defined as the in-plane vector obtained by rotating $\mathbf{e}_i$ clockwise by $90^\circ$, i.e. $\mathbf{e}_i^{*} \perp \mathbf{e}_i$. 
Let $\mathcal{T}_1$ and $\mathcal{T}_2$ be two adjacent triangles sharing the common edge $\mathbf{e}_i$, and let $\varphi$ denote the dihedral angle between their unit normals $\mathbf{n}_1$ and $\mathbf{n}_2$. 
The dihedral-angle-based representation of the discrete second fundamental form associated with edge $\mathbf{e}_i$ is written as
\begin{equation}
\mathbb{B}_{i}^{\mathrm{dih}}
=
\frac{\sin(\varphi/2)}{a_i/2}
\,
\left(
\bar{\mathbf e}_i^{*}
\otimes
\bar{\mathbf e}_i^{*}
\right),
\label{eq:dual-edge sff}
\end{equation}
where $\bar{\mathbf e}_i^{*} = \mathbf e_i^{*}/\|\mathbf e_i^{*}\|$ is the normalized dual edge. 
The geometric factor $a_i = 2A_{\mathcal{T}}/\|\mathbf e_i\|$ denotes the altitude of the triangle relative to the common edge, with $A_{\mathcal{T}}$ being the triangle area.
Here, the discrete normal field is defined by averaging the adjacent face normals on the edge, so that the variation of the normal within each triangle corresponds to $\varphi/2$.
The equivalence of the primal-edge (Equation~\eqref{eq:primal-edge sff}) and dual-edge (Equation~\eqref{eq:dual-edge sff}) representations has been rigorously established in \cite{clarisse2012discrete}. 
The virtual fold angle $\varphi_v$ defines a stress-free bending configuration. 
Evaluating the dihedral-based curvature at $\varphi_v$ yields the reference second fundamental form $\bar{\mathbb B}_i^{\mathrm{dih}} = \mathbb B_i^{\mathrm{dih}}(\varphi_v)$. 
The curvature deviation associated with the crease is therefore expressed as
\begin{equation}
\mathbb B_i^{\mathrm{dih}} - \bar{\mathbb B}_i^{\mathrm{dih}}
=
\frac{\sin(\varphi/2-\varphi_v/2)}{a_i/2}
\,
\left(
\bar{\mathbf e}_i^{*}
\otimes
\bar{\mathbf e}_i^{*}
\right).
\end{equation}
The difference $\varphi - \varphi_v$ therefore measures the deviation between the smooth geometric representation and the mesh-based geometry in the vicinity of the crease. 
This enables a direct geometric coupling between the bending measure and the crease rotation.
Following the standard shell kinematics, this curvature difference induces an additional bending strain contribution
\begin{equation}
\boldsymbol{\kappa}_i^{\mathrm{add}}
=
\bar{\mathbb A}_{ijk}^{-1}
\left(
\mathbb B_i^{\mathrm{dih}} - \bar{\mathbb B}_i^{\mathrm{dih}}
\right),
\end{equation}
where $\bar{\mathbb A}_{ijk}$ denotes the reference first fundamental form.
This additional bending strain is added to the conventional shell bending strain in Equation~\eqref{eq:bending strain} and contributes to the bending energy through Equation~\eqref{eq:bending energy}. 
From a continuum mechanics perspective, the additional bending contribution can be interpreted as replacing the natural (free) boundary condition of the shell by a prescribed rotational constraint along the crease, induced by the virtual fold angle $\varphi_v$.

\subsection{Dynamic Formulation for Active Origami Systems}

In addition to the internal elastic response arising from membrane, bending, and folding energies, the compliant origami robot is subjected to external environmental forces that are essential for reproducing realistic operational conditions. 
In this work, four types of external effects are incorporated into the simulation framework: gravitational loading $\mathbf{f}_{\mathrm{g}}$, normal contact force $\mathbf{f}_{\mathrm{contact}}^{N}$, tangential friction force $\mathbf{f}_{\mathrm{contact}}^{T}$, and magnetic braking $\mathbf{f}_{\mathrm{mag}}$. 
Within the proposed dynamic framework, all external contributions are assembled into a unified force vector 
$\mathbf{f}_{\mathrm{ext}}$ entering the governing equations of motion, i.e.,
\begin{equation}
\mathbf{f}_{\mathrm{ext}}
=
\mathbf{f}_{\mathrm{g}}
+
\mathbf{f}_{\mathrm{contact}}^{N}
+
\mathbf{f}_{\mathrm{contact}}^{T}
+
\mathbf{f}_{\mathrm{mag}}.
\end{equation}
It is worth noting that these environmental forces are applied only to physical vertices or panels and do not interact with the virtual crease elements. Consequently, the external force vector $\mathbf{f}_{\mathrm{ext}}(t)$ contains zeros in its last $N_{2}$ entries, corresponding to the degrees of freedom associated with the crease variables.

\paragraph{Gravity.} Gravity is modeled as a uniform body force acting on the mass of the origami sheet. 
Let $\rho$ denote the surface mass density and $h$ the thickness of the sheet. 
The gravitational acceleration vector is denoted by $\mathbf{g} \in \mathbb{R}^3$.
At the continuum level, the gravitational force density per unit area reads $\mathbf{f}_g = \rho h \mathbf{g}$.
After spatial discretization, the body force is converted into equivalent nodal forces 
consistent with the adopted mass formulation. 
Using a lumped mass representation, the gravitational force associated with vertex $a$ becomes
\begin{equation}
\mathbf{f}_{g}^{(a)} = m_{(a)} \mathbf{g},
\end{equation}
where $m_{(a)}$ is the lumped mass assigned to vertex $a$.
The global gravitational contribution to the external force vector 
$\mathbf{f}_{\mathrm{ext}}$ is therefore assembled as
\begin{equation}
\mathbf{f}_g
=
\left[
m_{(1)}\mathbf{g},
\dots,
m_{(N_{1})}\mathbf{g},
0,
\dots,
0
\right]^{\mathsf T}.
\end{equation}

\paragraph{Normal Contact.}
Normal contact is introduced to prevent interpenetration between the origami sheet and a rigid obstacle, while enabling momentum exchange during impact events.
Let $d$ denote the signed gap distance between node $a$ and the obstacle, and let $\tilde{d}$  be the prescribed contact threshold.
The outward unit normal on the obstacle is denoted by $\mathbf{n} \in \mathbb{R}^3$.
Contact is activated when
$d \le \tilde{d}$.
At the discrete level, the interaction is described by a nodal penalty potential
\begin{equation}
\Psi_{\mathrm{contact}}^{N}(d)
=
k_{\mathrm{contact}}^{N} \, (d - \tilde{d})^2
\ln\!\left(\frac{d}{\tilde{d}}\right),
\qquad d \le \tilde{d},
\end{equation}
where $k_{\mathrm{contact}}^{N}$ is the normal contact stiffness parameter.
And when $d>\tilde{d}$,  $\Psi_{\mathrm{contact}}^{N}(d) = 0$.
The corresponding normal contact force is obtained from the potential gradient as 
\begin{equation}
\mathbf{f}_{\mathrm{contact}}^{N}
=
- \frac{\partial \Psi_{\mathrm{contact}}^{N}}{\partial d}
\, \mathbf{n}.
\end{equation}
For implicit time integration, the consistent tangent contribution along the normal direction reads
\begin{equation}
\mathbb{J}_{\mathrm{contact}}^{N}
=
\frac{\partial^2 \Psi_{\mathrm{contact}}^{N}}{\partial d^2}
\left( \mathbf{n} \otimes \mathbf{n} \right).
\end{equation}
After assembly over all active nodes, the global normal contact contribution is incorporated into the external force vector
$\mathbf{f}_{\mathrm{ext}}$.

\paragraph{Tangential Contact.}
Tangential contact is introduced to account for frictional interaction along the contact interface.
It is activated only when normal contact is active, i.e., when $d \le \tilde{d}$.
At the discrete level, the frictional response follows a regularized Coulomb-type law.
Let
$N
=
\left\|
\mathbf{f}_{\mathrm{contact}}^{N}
\right\|$
denote the magnitude of the normal contact force, and let
$\dot{\mathbf{v}}$ be the nodal velocity with scalar speed
$v = \|\dot{\mathbf{v}}\|$. 
The friction coefficient is denoted by $\mu$.
The tangential contact force is defined in the direction of motion as
\begin{equation}
\mathbf{f}_{\mathrm{contact}}^{T}
=
- \mu \, N \,
f_{\mathrm{frc}}(v)\,
\frac{\dot{\mathbf{v}}}{\|\dot{\mathbf{v}}\|},
\qquad d \le \tilde{d}.
\end{equation}
Here $f_{\mathrm{frc}}(v)$ is a velocity-dependent regularization function introduced to ensure smooth behavior near zero velocity.
A typical $C^1$-continuous choice reads
\begin{equation}
f_{\mathrm{frc}}(v)
=
\begin{cases}
1,
& v \ge \varepsilon_v, \\[6pt]
- \left( \dfrac{v}{\varepsilon_v} \right)^2
+
 \dfrac{v}{\varepsilon_v} ,
& 0 \le v < \varepsilon_v,
\end{cases}
\end{equation}
where $\varepsilon_v$ is a small regularization parameter controlling the transition to full sliding friction.

\paragraph{Magnetic Actuation.}
Magnetic actuation models the interaction between magnetized faces of the origami sheet and an externally applied magnetic field\cite{sim2026electromagnetic,wang2023physicsaware}.
Let $\mathbf{B}_{\mathrm{ext}} \in \mathbb{R}^3$ denote a prescribed uniform external magnetic field.
Each magnetic triangular element 
$\mathcal{T}$ is characterized by a remanent magnetic flux density $\mathbf{B}_{r,(\mathcal{T})}$ defined over its current configuration.
The magnetic potential energy associated with element $\mathcal{T}$ is given by
\begin{equation}
\Psi_{\mathrm{mag},(\mathcal{T})}
=
\int_{A_{(\mathcal{T})}}
\mathbf{B}_{r,(\mathcal{T})} \cdot \mathbf{B}_{\mathrm{ext}}
\, \mathrm{d}A.
\end{equation}
Assuming a constant field over each element, this reduces to
\begin{equation}
\Psi_{\mathrm{mag},(\mathcal{T})}
=
\left(
\mathbf{B}_{r,(\mathcal{T})} \cdot \mathbf{B}_{\mathrm{ext}}
\right)
A_{(\mathcal{T})},
\end{equation}
where $A_{(\mathcal{T})}$ denotes the area of triangle.
The total magnetic potential energy reads
\begin{equation}
\Psi_{\mathrm{mag}}
=
\sum_{\mathcal{T}}
\Psi_{\mathrm{mag}}^{(\mathcal{T})}.
\end{equation}
The corresponding magnetic force and consistent tangent operator follow from the energy derivatives with respect to the nodal coordinates $\mathbf{v}$, 
\begin{equation}
\mathbf{f}_{\mathrm{mag}}
=
- \frac{\partial \Psi_{\mathrm{mag}}}{\partial \mathbf{v}},
\qquad
\mathbb{J}_{\mathrm{mag}}
=
\frac{\partial \mathbf{f}_{\mathrm{mag}}}{\partial \mathbf{v}}.
\end{equation}
After assembly over all magnetic faces, the global magnetic contribution is incorporated into the external force vector $\mathbf{f}_{\mathrm{ext}}$.

By unifying kinematic folding, elastic deformation, and interactions with the external environment within a single formulation, the framework enables systematic investigation of a wide range of mechanical behaviors central to origami-inspired robotic systems, including nonlinear dynamics\cite{gu2024kirigamiinspired}, multistability\cite{xi2023multistability,zhang2022kirigamibased}, snap-through transitions\cite{lu2026mechanical}, frictional contact\cite{jia2026multimode}, and multi-physics coupling\cite{xue2025origami}.

\section{Programmable Stability in Compliant Origami Units}
\label{sec:XX}

In compliant origami systems, mechanical response is no longer governed purely by kinematics, but by a delicate competition between panel bending, crease rotation, and geometric constraints.
This energy interplay gives rise to nonlinear stiffness variation, deformation mode selection, and bifurcation phenomena that are absent in purely rigid models.
Within the proposed DDG framework, crease and panel stiffness enter as explicit and independently tunable parameters.
Stability characteristics therefore become programmable structural properties rather than fixed geometric outcomes.

In this section, we systematically investigate how the interplay between crease rotational energy and panel bending energy shapes the nonlinear mechanical response of representative origami units.
Rather than serving solely as model validation, these studies reveal how stability characteristics—including deformation mode selection and symmetry-breaking transitions—can be continuously tuned across stiffness regimes.
We begin with a minimal single-fold (Z-fold) configuration under tensile loading to isolate bending–crease coupling mechanisms; and then examine shear-induced nonlinear response and bifurcation behavior, demonstrating how stiffness-controlled energy competition governs structural transitions.

\subsection{ Tensile Loading: Bending–Crease Coupling}

\begin{figure}[t]
    \centering
    \includegraphics[
        width=0.9\linewidth,
        trim={0 7cm 0 0},
        clip
    ]{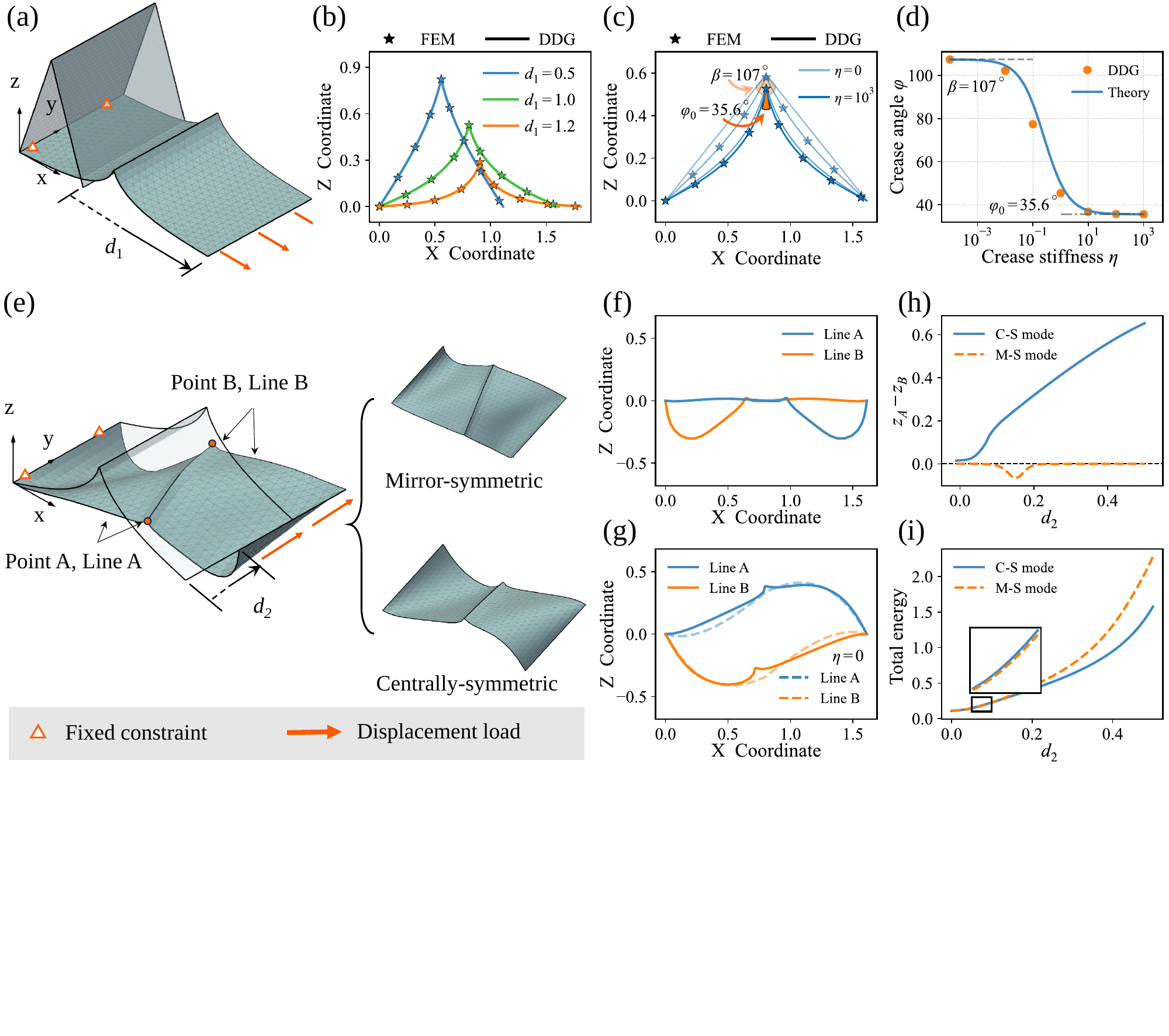}
    \caption{Mechanical response of the single-fold structure.
(a) Deformation configuration under tensile displacement loading.
(b,c) Boundary profile evolution for varying applied displacement $d_1$ and different crease stiffness values.
(d) Variation of the folding angle as a function of crease stiffness.
(e) Two distinct deformation modes under shear displacement loading: mirror-symmetric and centrally-symmetric configurations.
(f,g) Deformed boundary profiles corresponding to the two modes.
(h,i) Evolution of nodal displacement and total energy in the two modes.
}
    \label{fig:single_fold}
\end{figure}

As shown in {Figure~\ref{fig:single_fold}} a, a single-fold structure consists of two deformable panels connected along a shared edge, where a folding crease is defined.
In the discrete setting, the crease is represented by a sequence of identical rotational elements associated with the mesh, such that elastic deformation can occur not only within the panels but also along the crease itself.
The structure is loaded in tension by prescribing displacements in the x-direction at one end. Deformation profiles at loading lengths of $d_1$=0.5, 1.0, and 1.2 are extracted and compared with FEM results under identical loading conditions. 
The results in Figure~\ref{fig:single_fold} b demonstrate a high level of consistency between the DDG-based predictions and the FEM simulations.

By systematically varying the ratio between hinge stiffness and panel bending stiffness, this example enables us to investigate how the proposed DDG framework continuously transitions between rigid origami behavior and flexible deformation. 
Figure~\ref{fig:single_fold} c shows the extracted deformation profiles of the single-fold structure for different values of the crease stiffness.
To characterize the coupled effect of crease rotation and panel bending on the resulting origami shape, we introduce a dimensionless stiffness ratio
\begin{equation}
\eta = \frac{K_{\mathrm{r}} W}{E h^3},
\end{equation}
where $K_{\mathrm{r}}$ denotes the rotational stiffness density of the discrete crease, $W$ is a characteristic length of the origami element, it is approximately the size of the panel perpendicular to the hinge, and $E$ and $h$ are the Young’s modulus and thickness of the panels, respectively.
For $\eta \ll 1$, corresponding to relatively low crease stiffness, deformation localizes at the crease and the mechanical response approaches the rigid origami limit.
In contrast, when $\eta \gg 1$, which means the crease stiffness dominates over panel bending stiffness, the structure deforms primarily through distributed panel curvature.
In both regimes, the deformation profiles predicted by the DDG framework exhibit excellent agreement with FEA.

This transition is further quantified by examining the fold angle as a function of $\eta$, as shown in Figure ~\ref{fig:single_fold} d.
The response exhibits two distinct plateau regions.
For small $\eta$, the fold angle is dictated purely by geometric compatibility, corresponding to a rigid kinematic folding mode, we record the current fold angle as $\beta = 107^\circ$.
In contrast, for large $\eta$, the fold angle approaches its initial value $\varphi_0 = 35.6^\circ$, indicating that rotation at the crease is suppressed and deformation is accommodated primarily through panel bending.
Between these two limits, both crease rotation and panel bending contribute to the deformation, resulting in a smooth transition between the two plateaus.

To capture this energy competition, a simplified energy-based Ritz model based on a Ritz approximation is developed, with the detailed derivation provided in Appendix~A.
The resulting closed-form expression for the fold angle is
\begin{equation}
\varphi = \beta + \frac{\eta(\varphi_0 - \beta)}{1/2 + \eta},
\end{equation}
which explicitly reveals the balance between crease stiffness and panel bending rigidity.
In the limit $\eta \to 0$, the fold angle reduces to $\varphi = \beta$, corresponding to pure kinematic folding.
Conversely, as $\eta \to \infty$, the fold angle approaches $\varphi = \varphi_0$, indicating that the crease remains near its initial configuration.
As shown in Figure ~\ref{fig:single_fold} d, the theoretical prediction agrees well with the numerical results. 
This simple example demonstrates that the proposed DDG framework not only recovers rigid origami behavior as a limiting case, but also naturally captures flexible deformation and its governing mechanics within a unified formulation.
These features form the foundation for the analysis of more complex origami structures presented in the following sections.

\subsection{Shear Loading: Nonlinear Response and Bifurcation}
Further case studies demonstrate the capability of the proposed framework in capturing nonlinear elastic behavior.
As shown in Figure ~\ref{fig:single_fold} e, after the initial tensile loading, additional displacements $d_2 $ are applied in the y-direction.
Under the combined action of bending and shear, the structure exhibits two distinct stable post-buckling bifurcation modes: a mirror-symmetric mode and a centrally-symmetric mode.

By extracting the deformation profiles along the two edges highlighted in blue and orange, Figures~\ref{fig:single_fold}(f) and (g) illustrate the characteristic deformation patterns associated with the two deformation modes, respectively.
In the mirror-symmetric mode, the deformation profiles of the two edges are mirror images of each other with respect to the central axis in the $x$-direction.
In contrast, in the centrally symmetric mode, the two edge profiles are related by point symmetry about the structural center.
Figure~\ref{fig:single_fold}(h) plots the difference in the $z$-coordinates of two points, A and B, located at the crease, as a function of the applied loading during deformation.
In the mirror-symmetric mode, the two points generally remain at the same height throughout the loading process.
By contrast, in the centrally symmetric mode, the height difference between the two points gradually increases with loading.
The bifurcation between the two modes occurs from the very onset of the $d_2$ loading.

The activation of these two modes depends on two key factors: the presence of crease rotational stiffness and the existence of initial perturbations.
In the absence of crease stiffness, the centrally-symmetric mode emerges as the unique stable deformation path\cite{shi2025double}.
When finite crease stiffness is introduced, the mirror-symmetric mode arises spontaneously as a preferred deformation pattern.
This behavior can be attributed to the constraint imposed by the crease, which suppresses bending deformation orthogonal to the crease direction and thereby inhibits the spontaneous development of the centrally-symmetric mode.

Nevertheless, the mirror-symmetric mode is not globally stable.
Figure~\ref{fig:single_fold}(i) compares the energy evolution associated with the two deformation modes.
For most of the loading process, the centrally symmetric mode exhibits a lower total energy and therefore represents the energetically preferred configuration.
However, during a short initial stage of loading, the mirror-symmetric mode temporarily possesses lower energy, which explains why it can be naturally triggered in the absence of external perturbations.
When a small perturbation is introduced at the crease, for example, breaking the symmetry, the structure ultimately transitions to the centrally symmetric mode, indicating that this mode corresponds to the globally stable post-buckling configuration.

Together, these results demonstrate that stability characteristics can be continuously tuned through crease stiffness modulation.
This stiffness-dependent tunability provides the mechanical foundation for the programmable robotic behaviors explored in subsequent sections.

\section{Bridging Rigid and Compliant Origami Regimes}

The design of compliant origami robots often involves structures that operate across a wide range of material flexibility, from nearly rigid panels to highly compliant membranes. 
Accurately predicting their mechanical response requires a modeling framework capable of handling both rigid folding and large-deformation compliant behavior within a single formulation. Miura origami provides an ideal testbed for this purpose\cite{callens2018flat}.

Although governed by a single geometric pattern, Miura structures have been realized as rigid kinematic metamaterials\cite{schenk2013geometry}, moderately flexible systems for programmable surface shaping\cite{dudte2016programminga}, and highly compliant membranes for deployable aerospace applications\cite{wang2024deployment}. 
Across these realizations, variations in structural flexibility give rise to fundamentally different deformation mechanisms.
In this section, we investigate Miura origami across this stiffness spectrum through three representative cases: (i) rigid kinematic folding, (ii) out-of-plane bending in moderately flexible configurations, and (iii) large-deformation deployment of highly compliant membranes. 
Together, these studies demonstrate the capability of the DDG framework to consistently model rigid and compliant behaviors within a single formulation.

\begin{figure}[t]
    \centering
    \includegraphics[
        width=0.9\linewidth,
        trim={0 0 0 0},
        clip
    ]{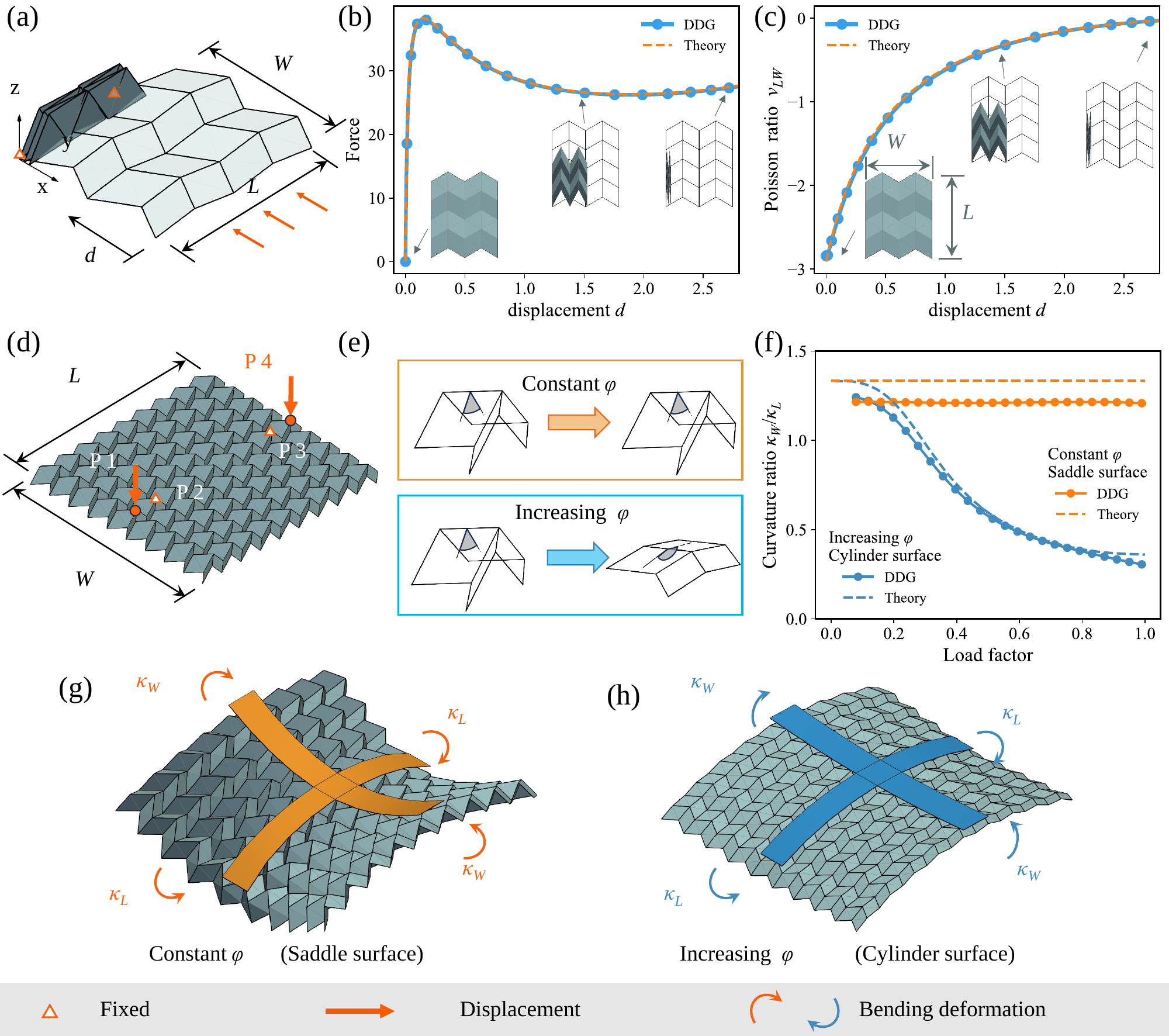}
    \caption{Mechanical behavior of rigid and flexible Miura origami structures.
(a) Compressive deformation of the rigid Miura configuration, and 
(b) corresponding load--displacement response.
(c) Evolution of the effective in-plane Poisson's ratio. 
(d) Four-point bending configuration applied to the flexible Miura structure. The theoretical model is derived from Ref. \cite{wei2013geometric,schenk2013geometry}.
(e) Evolution of the folding angle.
(f) Comparison between the simulated curvature ratio and the analytical prediction.
(g,h) Two deformed configurations under bending, exhibiting saddle-shaped and cylindrical surface morphologies.
}
    \label{fig:Miura_Origami}
\end{figure}

\subsection{Kinematic Folding in the Rigid Limit}
As illustrated in {Figure ~\ref{fig:Miura_Origami}} (a), we begin with the rigid origami limit by considering a structure composed of four fundamental Miura unit cells, where the dimensionless stiffness ratio $\eta$ is small, indicating that the crease stiffness is much small compared to the panel bending stiffness.
Each Miura unit consists of a single four-valent vertex formed by three mountain folds and one valley fold, connecting four parallelogram facets.
A key feature of this design is its rigid-foldability, which enables large, coordinated folding motions through variations in fold angles while preserving facet rigidity. 

A quasi-static compression is applied to the rigid Miura structure along the $x$-direction.
As shown in Figure ~\ref{fig:Miura_Origami} (a), the left end of the structure ($x=0$) is constrained in the $x$-direction, while a prescribed displacement is applied to all nodes at the right end toward the negative $x$-direction.
Starting from an almost flat configuration, the reaction forces at the right end are recorded during the compression process.
The total reaction force in the $x$-direction is plotted against the applied displacement in Figure ~\ref{fig:Miura_Origami} (b), yielding the load–displacement response of the structure.
%
The slope of the load–displacement curve reflects the effective axial stiffness of the Miura structure.
Following an initial regime of high stiffness, the structure rapidly transitions into a negative-stiffness regime, after which the stiffness slightly increases as folding progresses.
We compare the numerical results obtained from the proposed DDG framework with analytical kinematic predictions  \cite{wei2013geometric,schenk2013geometry}, and excellent agreement is observed, validating the accuracy of the numerical model in capturing rigid origami behavior.

In addition to its folding kinematics, the Miura structure is well known for its negative Poisson’s ratio, which means that the Miura origami contracts simultaneously in both the $x$- and $y$-directions.
The numerical simulations accurately reproduce this behavior, as shown in Figure ~\ref{fig:Miura_Origami} (c).

\subsection{Elastic Curvature Programming in Moderately Flexible Miura} 

Kinematic analyses of rigid Miura origami indicate that its deformation is restricted to in-plane stretching and compression modes. 
However, thin sheet materials such as paper are inherently flexible and readily undergo bending.
The introduction of bending compliance endows Miura origami with additional deformation degrees of freedom, enabling out-of-plane bending behaviors beyond the rigid-folding limit.

Previous theoretical work demonstrated that when a Miura structure is subjected to a uniaxial bending moment, it simultaneously bends in two orthogonal directions \cite{wei2013geometric}. 
This coupling between orthogonal curvatures is referred to as the \emph{bending Poisson effect}. 
To investigate this phenomenon, we apply a four-point bending configuration to the Miura structure shown in Figure ~\ref{fig:Miura_Origami} (d). 
Displacement constraints in the $z$-direction  are imposed at points $P_2$ and $P_3$ near the edges along the symmetry axis in the $L$ direction, while prescribed downward displacements are applied at points $P_1$ and $P_4$ located on the boundary of the same line. 
To allow bending deformation, the panel bending stiffness is appropriately reduced.

A representative simulation result is shown in  Figure~\ref{fig:Miura_Origami}(g).
The structure bends along the loading direction $L$ while simultaneously exhibiting bending of opposite sign along the transverse direction $W$, forming a saddle-shaped surface. 
The bending Poisson ratio is defined as the negative ratio between the principal curvatures in the orthogonal and loading directions, i.e. $\nu_b = -\kappa_W / \kappa_L$.
Geometric analysis predicts that this quantity is equal to the negative in-plane Poisson ratio of the Miura pattern\cite{wei2013geometric}, 
\begin{equation}
-\kappa_W/\kappa_L = -\nu_{WL} .
\label{eq:bend Possion}
\end{equation}
Due to the negative Poisson ratio of Miura origami, $\kappa_W$ and $\kappa_L$ always have opposite signs, implying that the bent configuration is intrinsically saddle-shaped.

Beyond this classical response, our simulations reveal a more complex deformation mode in which the Miura structure can evolve into an approximately cylindrical surface, as shown in Figure ~\ref{fig:Miura_Origami} (h). 
The key distinction between the saddle-shaped configuration Figure ~\ref{fig:Miura_Origami} (g) and the cylindrical one Figure ~\ref{fig:Miura_Origami} (h) lies in whether the fold angle $\varphi$ undergoes significant variation during deformation. 
Accordingly, we distinguish two deformation regimes: a \emph{constant-$\varphi$} regime and an \emph{increasing-$\varphi$} regime, as illustrated in Figure ~\ref{fig:Miura_Origami} (e).

Building on the analysis in the previous section, when the crease stiffness is sufficiently large, the structure favors panel bending while maintaining nearly constant fold angles, leading to a saddle-shaped geometry. 
In contrast, for smaller crease stiffness, deformation is dominated by fold rotation, resulting in a substantial increase in $\varphi$ and an overall cylindrical bending mode. 
Figure ~\ref{fig:Miura_Origami} (f) plots the evolution of the bending Poisson ratio during the bending process for both regimes. 
In the constant-$\varphi$ case, the bending Poisson ratio remains nearly constant and approaches the theoretical prediction $-\kappa_W/\kappa_L \approx 1.3$. 
In the increasing-$\varphi$ case, as the structure progressively unfolds and approaches a nearly flat configuration, the bending Poisson ratio decreases and approaches $1/3$. 
At this stage, the deformation becomes predominantly cylindrical.

It is important to emphasize that directly applying the closed-form theoretical prediction derived under the constant-$\varphi$ assumption to the increasing-$\varphi$ regime inevitably leads to discrepancies. 
This discrepancy is not caused by a breakdown of the geometric relation, but by the fact that the fold angle evolves during deformation and cannot be treated as constant.
When the instantaneous fold angles $\varphi$ extracted from the simulations are substituted into the analytical expression for the bending Poisson ratio, the theoretical predictions recover excellent agreement with the numerical results. 
This observation demonstrates that the fundamental geometric relation in Equation ~\eqref{eq:bend Possion} remains valid throughout the deformation process. 
Crucially, however, the effective Poisson ratio $\nu_{WL}$ must be evaluated as a function of the evolving fold angle $\varphi$, rather than treated as a constant.

Under large bending deformations, the Miura array develops spatially nonuniform curvature distributions, and the fold angles vary both temporally and spatially across the structure.
These nonlinear effects limit the direct applicability of closed-form solutions based on fixed geometric parameters\cite{liu2017nonlinear}. 
The numerical framework proposed in this work therefore provides a consistent means of tracking the evolution of fold angles and curvature fields, allowing accurate prediction of the global behavior of Miura origami.

\subsection{Highly Flexible Deployable Membrane Behavior}

Miura origami structures are widely used in aerospace applications due to their excellent deployability, particularly for ultra-thin membrane systems requiring in-orbit deployment.
Such structures combine high in-plane flexibility with crease constraints, resulting in strongly nonlinear responses during deployment\cite{wang2023simulation}.
The crease–membrane coupled framework developed in this work is therefore well suited for simulating the deployment behavior of highly flexible Miura membranes.
To reflect representative deployment strategies, both displacement-driven and force-driven loading schemes are considered, corresponding to cable-driven and thruster-driven deployment, respectively.
In the numerical model, loads are applied at the four corner points P1 - P4, as shown in {Figure ~\ref{fig:Miura_mem}} (a). 

\begin{figure}[t]
    \centering
    \includegraphics[
        width=0.9\linewidth,
        trim={0 15cm 0 0},
        clip
    ]{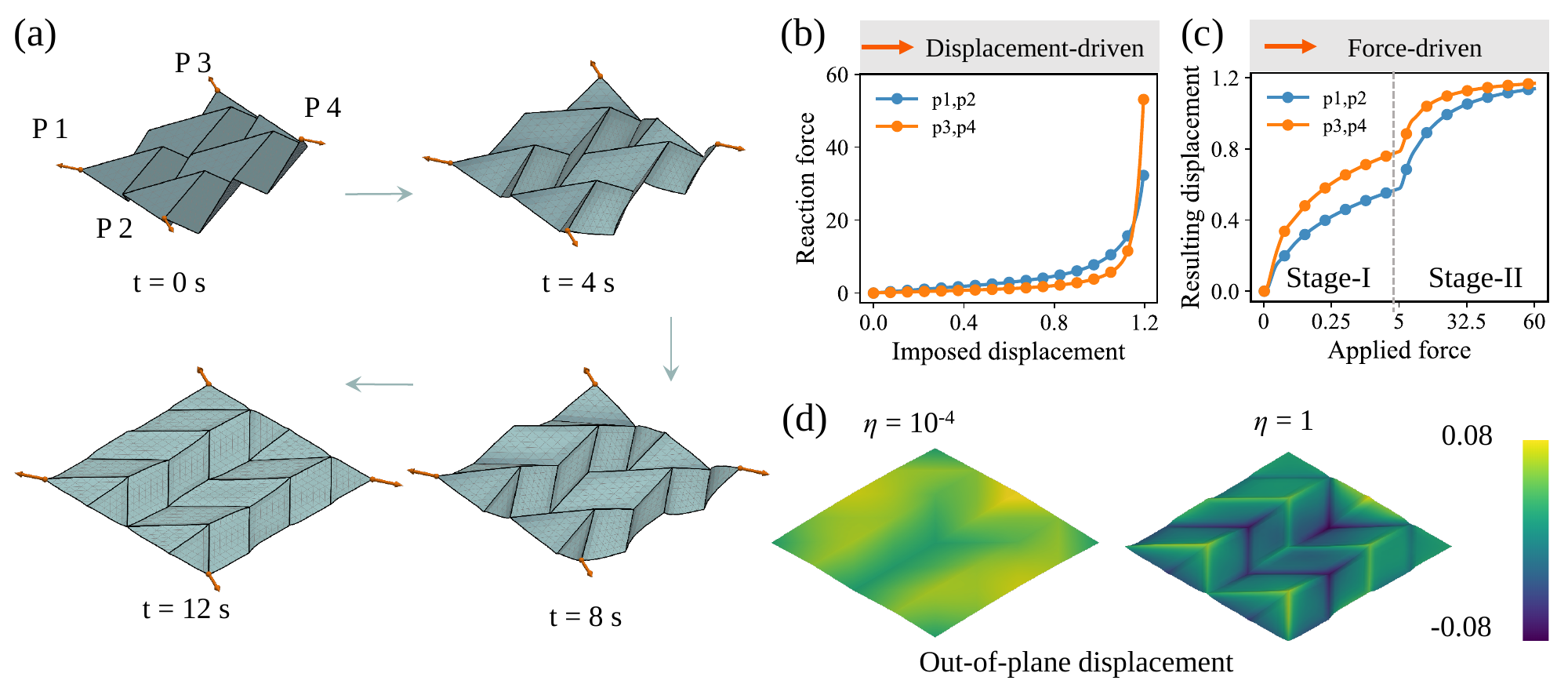}
    \caption{Deployment simulation of Miura origami membranes.
(a) Evolution of the deployed configuration at representative time instants. The deployable membrane configuration is derived from Ref. \cite{wang2023simulation}.
(b,c) Load--response curves under displacement-controlled and force-controlled actuation, respectively.
(d) Influence of crease stiffness on the surface flatness of the deployed membrane.
}
    \label{fig:Miura_mem}
\end{figure}

Figure ~\ref{fig:Miura_mem} (a) illustrates the deployment process under displacement-driven loading, in which constant-velocity displacement constraints of $0.1/s$ are applied at the four corner points.
The loading direction at each corner is defined by the vector from its initial position to the corresponding position in the fully deployed configuration. The resulting imposed displacements and reaction forces are shown in Figure ~\ref{fig:Miura_mem} (b).
Owing to the symmetry of both the structure and the loading scheme, the reaction forces at P 1 and 2 are equal in magnitude and opposite in direction, and the same symmetry holds for P 3 and 4.
The load–displacement response clearly reveals two distinct mechanical stages during the deployment process.
In the first stage (approximately $0s -8s$), deployment is primarily accommodated by bending deformation of the membrane panels, leading to a relatively low structural stiffness and a gradual increase in the reaction force.
In the second stage, further deployment is dominated by changes in the fold angles as the structure approaches a fully flattened configuration.
As the overall geometry becomes increasingly planar, the moment arm associated with the external loads diminishes, and substantially larger forces are required to drive additional crease rotation, resulting in a pronounced stiffening response.

Under force-controlled loading (Figure ~\ref{fig:Miura_mem} (c)), a staged loading strategy is adopted in view of the two-stage deployment behavior observed above.
Specifically, the applied forces at P1–P4 are equal in magnitude, and are gradually increased from $0$ to $5$ in the first stage and then further increased from $5$ to $60$ in the second stage. 
This loading protocol effectively mitigates excessive variations in the deployment rate at both the early and late stages, enabling a relatively stable evolution throughout the entire deployment process.
The numerical results demonstrate that the proposed framework can robustly capture the complex nonlinear mechanical response, thereby providing a reliable basis for the design and optimization of deployment actuation and control strategies.

Due to the finite stiffness of the creases, the final deployed configuration is not an ideal flat state, but is instead governed by the coupled effects of membrane bending and crease constraints.
Figure ~\ref{fig:Miura_mem} (d) compares the deployed configurations for two representative crease stiffness values, $\eta = 1\times10^{-4}$ and $\eta = 1$, where the color contours indicate the out-of-plane displacement of each point relative to the mean height.
It can be observed that, for a larger crease stiffness, the structure exhibits pronounced surface non-uniformity even after full deployment.
Such behavior cannot be captured by conventional deployment models that neglect crease stiffness, highlighting the necessity of crease--membrane coupled modeling for the accurate analysis of highly flexible deployable structures\cite{wang2024deployment}.

\section{From Structural Mechanics to Robotic Functionality}

Building upon the structural analyses presented above, we further extend the framework to origami-inspired compliant robotic systems, where mechanical response directly translates into functional motion. 
Two representative examples are considered: a jumping robot based on the bistable behavior of the Waterbomb unit, and a crawling robot driven by the directional actuation of a Kresling mechanism. 
In next two sections, we examine how these structural principles inform robotic design, including actuation strategies and motion generation, and demonstrate that the proposed DDG framework enables provides a consistent dynamic modeling framework for simulating complex robotic motion under nonlinear and multistable structural responses.

\subsection{Jumping Robot: Bistable Actuation in Waterbomb Origami}

The Waterbomb origami is a classical folding pattern characterized by an alternating arrangement of mountain and valley folds around a central vertex\cite{ma2020folding,feng2018twist}.
Owing to its geometric nonlinearity, this configuration is known to exhibit bistable behavior\cite{chen2016symmetric,hanna2015force}.
In this section, two numerical case studies are presented to investigate the bistability of Waterbomb structures, considering (i) a flexible waterbomb and (ii) a rigid waterbomb.
For both cases, the proposed framework successfully captures the existence of two stable equilibrium states, as well as the snap-through behavior associated with dynamic transitions between them.
These examples demonstrate the capability of the present approach to model bistable responses and complex nonlinear dynamics in origami structures across different stiffness regimes. 

\begin{figure}[t]
    \centering
    \includegraphics[
        width=0.9\linewidth,
        trim={0 0 0 0},
        clip
    ]{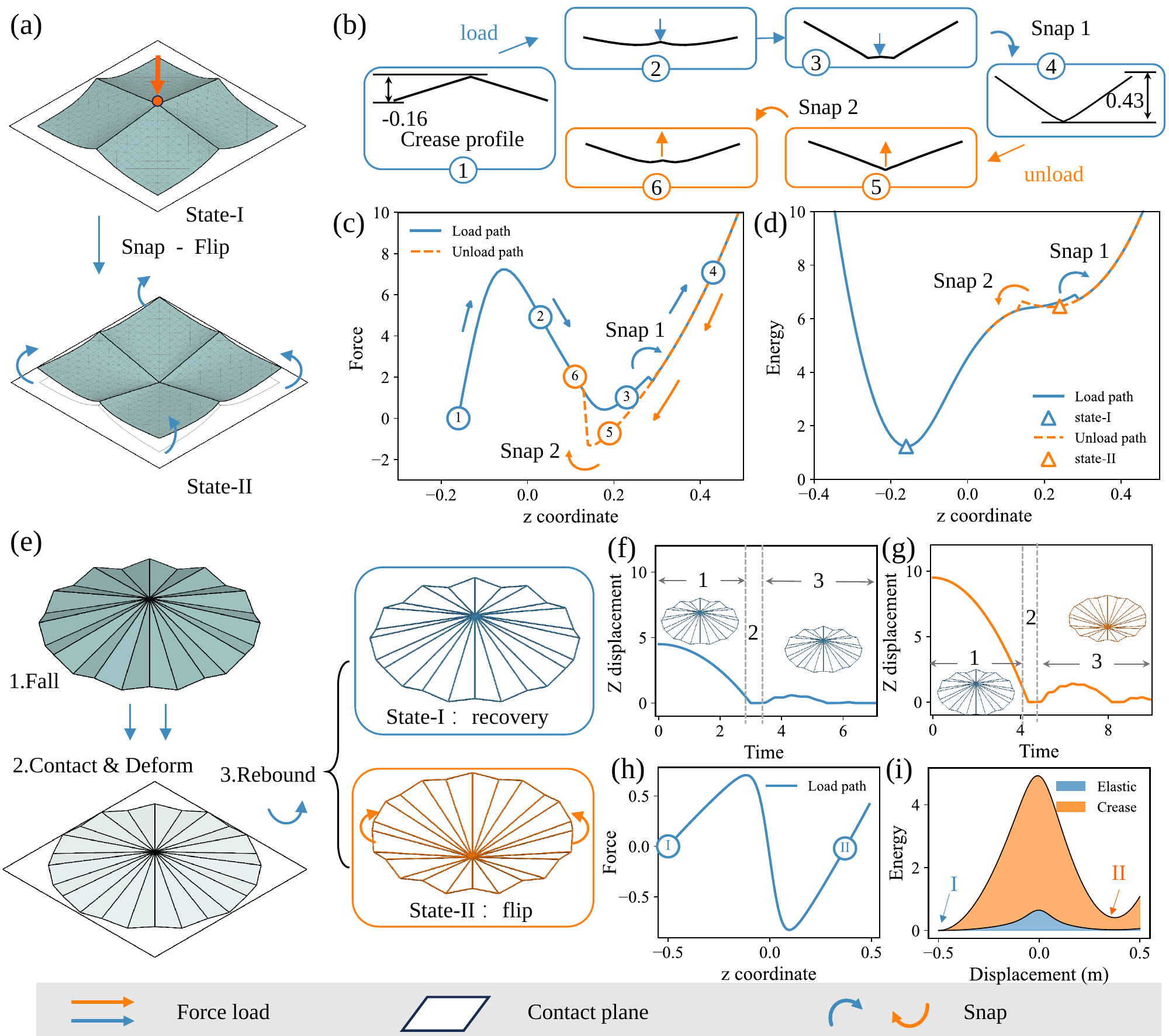}
    \caption{Bistable mechanical behavior of the Waterbomb origami.
(a) Two stable states of the flexible Waterbomb configuration and the associated flip.
(b) Deformation profiles at representative instants during the state-switching process.
(c,d) Load--displacement response and corresponding energy evolution.
(e) Two stable states of the rigid Waterbomb configuration\cite{ta2022printable} and its snap-through behavior.
(f,g) Time--displacement histories under free fall from heights $z=5$ and $z=10$, leading to recovery and flip, respectively.
(h,i) Quasi-static load--displacement response and energy distribution .
}
    \label{fig:Waterbomb}
\end{figure}

\subsubsection{ Bistable of flexible waterbomb}

The flexible case is constructed from a simple origami configuration.
Starting from a square sheet, the paper is folded along two symmetry axes parallel to the edges, resulting in four mountain creases with intrinsic fold angles induced by plastic deformation, as illustrated in {Figure ~\ref{fig:Waterbomb}} (a).
For a purely rigid origami model, a single unit composed solely of four mountain folds cannot achieve a geometrically compatible deformation, as this would require an isometric mapping from a planar surface to a configuration with nonzero Gaussian curvature, in violation of Gauss' Theorema Egregium\cite{callens2018flat}.
In contrast, when material flexibility is allowed, the geometric incompatibility introduced by the creases can be accommodated through bending deformation of the panels.
As shown in Figure ~\ref{fig:Waterbomb} (a) (state--I), spontaneous bending emerges in the central region enclosed by the four creases.
This bending deformation effectively plays the role of a virtual valley fold, and the resulting configuration is therefore identified as a flexible variant of the Waterbomb origami.
When the flexible Waterbomb is placed on a flat surface, pressing the central region induces an up--down inversion of the boundary, while the inverted configuration (state--II) remains stable after the external load is removed.
The coexistence of these two stable configurations demonstrates the bistable nature of the structure.

To numerically capture this bistability and the associated snap-through behavior, dynamic simulations are performed using the proposed framework.
The model consists of four flexible panels connected by four creases and interacting with a rigid horizontal plane.
Gravity is included, and a normal contact condition is enforced between the panels and the supporting plane to represent planar constraints during the inversion process.
The intrinsic fold angles are prescribed to represent the naturally formed creases.
As these intrinsic angles are applied, bending deformation develops in the panels, and the four corners act as stable supports, giving rise to the initial equilibrium configuration (state--I).
A downward force is then applied at the center of the structure, driving the central point toward the plane.
Under the combined effects of crease forces and panel bending, the structure undergoes inversion and transitions to the second stable equilibrium configuration (state--II).

To gain further insight into the structural response and equilibrium states, quasi-static simulations are performed by driving the central point of the structure along the vertical ($z$) direction.
The corresponding load--displacement response and energy evolution are obtained, and the profiles of the crease edges at six representative configurations are plotted in Figure ~\ref{fig:Waterbomb} (b).
The load--displacement curve is shown in  Figure ~\ref{fig:Waterbomb} (c), where a downward displacement load is applied at the center.

Along the loading path (blue curve), the structure initially exhibits a short regime of positive stiffness, followed by a negative stiffness region and subsequently a return to positive stiffness as the deformation proceeds from configuration~1 to~4.
During this process, the relative height between the center and the boundary undergoes a reversal.
Notably, the crease-edge profile at configuration~3 is not smooth, suggesting that this state corresponds to an unstable equilibrium.
As a result, the structure undergoes a small-amplitude snap-through transition from configuration~3 to~4, recovering stability.
Configuration~4 coincides with the second stable equilibrium (state--II).
This snap-through event is manifested as a discontinuous jump in the load--displacement curve and is denoted as snap~1.

After reaching the second stable state, the structure is unloaded, as indicated by the orange curve Figure ~\ref{fig:Waterbomb} (c).
Owing to the asymmetry between the two stable configurations, the unloading path does not retrace the loading path.
Instead, the structure deforms stably over a finite range and, while remaining in the state--II configuration, passes beyond $z = 0.2$.
After configuration~5, a second snap-through event (snap~2) occurs, driving the system back to the primary equilibrium branch.

The mechanism behind the two snap events is revealed by the energy curves in Figure ~\ref{fig:Waterbomb}(d).
The two stable configurations correspond to local minima on different energy branches: the local minimum on the loading branch (blue curve) defines stable state--I, whereas the local minimum on the unloading branch (orange curve) defines stable state--II.
Since these two energy paths do not coincide, stable state--II cannot be reached continuously along the loading branch.
Instead, once the system passes beyond the vicinity of state--II while remaining on the loading path, it resides in a higher-energy unstable configuration and subsequently relaxes onto the stable unloading branch, giving rise to snap~1.
An analogous mechanism governs the occurrence of snap~2 during unloading.

\subsubsection{ Jumping behavior of the rigid waterbomb}

The bistability of the Waterbomb structure enables energy storage during transitions between its two stable states, which can be exploited to realize a jump origami robot \cite{ta2022printable}.
To illustrate the capability of the proposed framework in capturing the dynamics of such a system, we perform numerical simulations based on the mechanism described in the literature.

The robot model consists of a rigid Waterbomb origami with $12$ mountain and valley creases, exhibiting two stable configurations: the initial configuration (state--I) and its inverted configuration (state--II), as shown in Figure ~\ref{fig:Waterbomb} (e).
When released from a certain height, the robot undergoes three sequential stages: free fall, deformation, and energy storage upon contact with the ground, and subsequent rebound.
Depending on the gravitational potential energy imparted during the fall, the robot may either recover its initial configuration (recovery state--I) or transition to the inverted configuration (state--II) after the rebound.
Figure ~\ref{fig:Waterbomb} (f) and (g) depict the trajectories of the Waterbomb robot released from heights of $z=5$ and $z=10$, respectively.
After completing the three stages, the robot released from $z=5$ does not acquire sufficient energy to overcome the energy barrier and returns to state--I, whereas the robot released from $z=10$ successfully transitions to state--II.

The load--response behavior of the rigid Waterbomb robot is shown in Figure ~\ref{fig:Waterbomb} (h).
The curve exhibits a sequence of positive stiffness, negative stiffness, and positive stiffness during the transition between stable states, similar to the corresponding response of the flexible robot (Figure ~\ref{fig:Waterbomb} (c)).
Compared with the flexible system, the rigid robot displays a smoother transition between stable states, avoiding the small-amplitude snap events observed in the flexible case.
The corresponding energy distribution during state transitions is illustrated in Figure ~\ref{fig:Waterbomb} (i), indicating that the deformation energy in the rigid origami is primarily concentrated in the creases.
By applying external forces to drive the robot over the energy barrier, the input energy is temporarily stored in the structural deformation and subsequently converted into kinetic energy, thereby enabling the jump mechanism\cite{ta2022printable}.

\subsection{Crawling Robot: Directional Actuation in Kresling Origami}

The Kresling origami pattern is a cylindrical shell structure exhibiting intrinsic multistability. 
Its geometric nodes are defined by the intersections of two families of longitudinal helical creases and one set of transverse circular creases. 
Owing to its axial stackability, the Kresling structure enables directional actuation through the assembly of multiple units.
Building upon this structural advantage, a compact magnetically actuated soft origami robot has been designed \cite{ze2022soft}. 
Through a clever geometric design and judicious magnetic torque distribution, they achieved controllable in-plane contraction and directional locomotion using a remarkably simple external-field actuation scheme. 
In this section, the proposed unified DDG-based framework is employed to simulate the complex nonlinear dynamics of this Kresling-based robotic system.

\subsubsection{ Multistability of the rigid kresling system}

For an individual Kresling unit, the fundamental folding and deployment motions can be achieved in two distinct ways: by applying an axial compressive displacement, or by prescribing a relative rotational angle between the upper and lower annular end faces, as illustrated in {Figure ~\ref{fig:kresling}} (a) and  (d).
Figure ~\ref{fig:kresling} (b) and (c) present the force--displacement response under axial loading and the corresponding energy evolution, respectively. 
As the compressive displacement of the end faces increases, the structural response exhibits a characteristic positive--negative--positive stiffness transition. 
This nonlinear behavior indicates bistability of the structure: the reaction force vanishes at both the fully deployed configuration ($d = 0$) and the fully folded configuration ($d \approx 0.4$), which correspond to two local minima in the total potential energy landscape.

\begin{figure}[t]
    \centering
    \includegraphics[
        width=0.9\linewidth,
        trim={0 0 0 0},
        clip
    ]{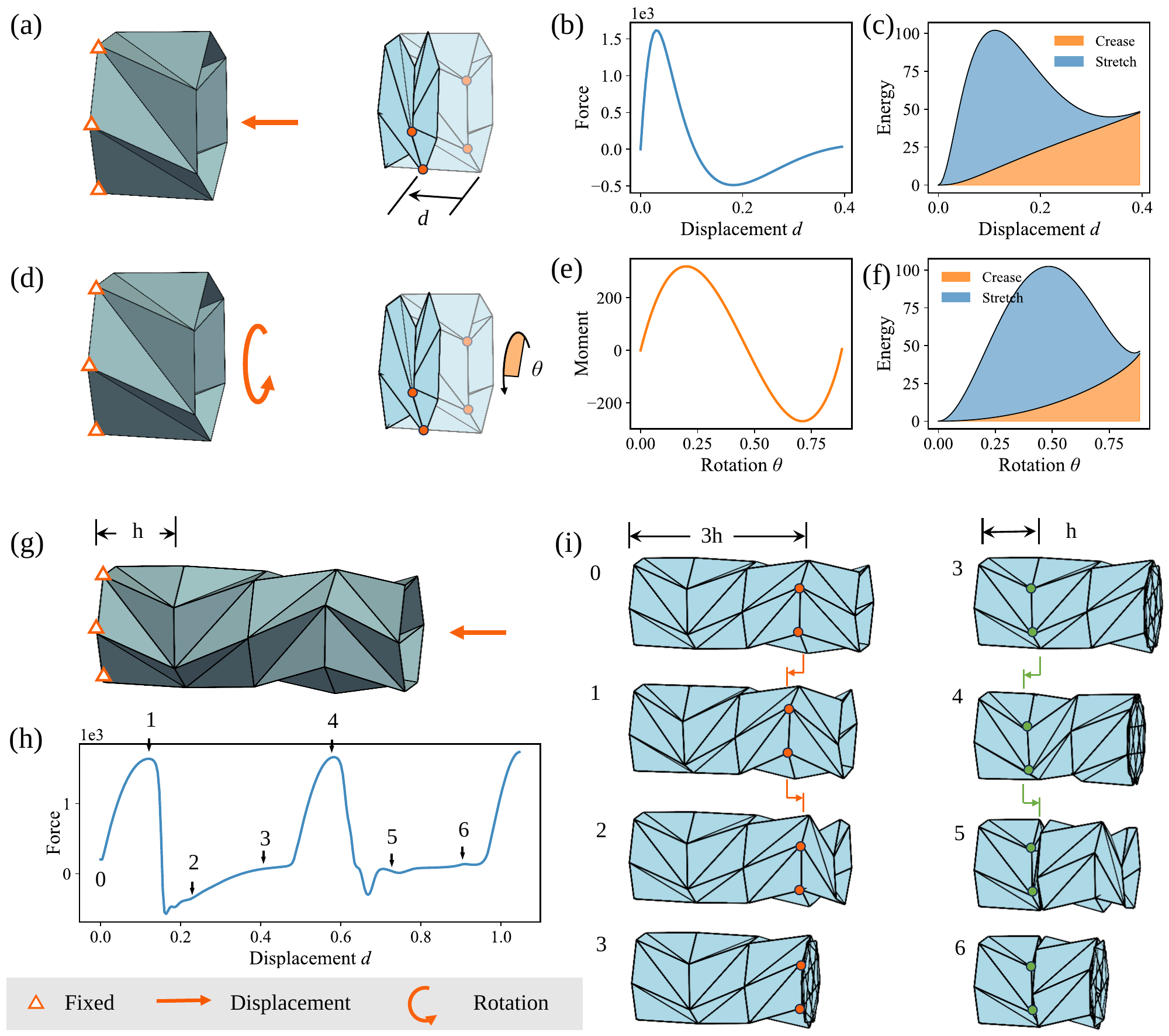}
    \caption{Multistable behavior of the rigid Kresling origami.
(a--c) Load--displacement response and corresponding energy landscape of a single rigid Kresling unit under axial displacement loading.
(d--f) Load--displacement response and energy distribution under circumferential rotational displacement loading.
(g) Dynamic switching simulation of a four-unit rigid Kresling assembly exhibiting multistable transitions.
(h) Time histories associated with the dynamic switching process.
(i) Deformed configurations at representative time instants during multistable transition.
}
    \label{fig:kresling}
\end{figure}

In contrast to the waterbomb structure, where bistable transitions are governed by the coupling between panel bending and crease rotation, the multistability of a Kresling unit primarily arises from panel stretching induced by variations in crease lengths. 
During the state-switching process, the structure must overcome a pronounced stretching energy barrier. 
The energy distribution clearly reveals this mechanism: as loading proceeds, the stretching energy increases rapidly, reaches a peak at the energy barrier, and subsequently decreases after the transition, dominating the overall energy evolution throughout the process.
Figure ~\ref{fig:kresling} (e) and (f) illustrate the mechanical response under circumferential torsional loading.
The torque-angle curve also exhibits bistable characteristics, and the physical mechanism is consistent with that in the axial compression case.

Four bistable Kresling units are stacked axially, each with an initial length of $h$, forming an integrated structure that exhibits more complex multistable behavior. 
Dynamic simulations are performed to capture the loading process of the stacked system. 
As shown in Figure ~\ref{fig:kresling} (g), the left end is fixed while a prescribed compressive displacement is applied at the right end. 
The multi-unit assembly displays pronounced nonlinear state-switching behavior during compression.

Figure ~\ref{fig:kresling} (h) shows the sum of the reaction forces of all nodes on the right end face as a function of the applied displacement. 
Six representative configurations are extracted in Figure ~\ref{fig:kresling} (i) to illustrate the transition process. 
Configurations 0--3 correspond to one complete stable-state transition, which can be divided into three stages:
\textbf{In stage 0--1,} the imposed displacement simultaneously compresses all four units, leading to axial shortening in each. 
The combined length of the bottom three units gradually decreases to below $3h$.
\textbf{In stage 1--2,} When the reaction force reaches the critical peak value ($F \approx 1.5 \times 10^{3}$), the bottom three units undergo snap-back instability and rapidly return to their original combined length of $3h$, while the outermost unit loses stiffness and is quickly compressed.
\textbf{In stage 2--3,} The outermost unit continues to compress until it becomes fully folded.

Subsequently in stage 3--6, the system repeats the aforementioned sequence of instability, recovery, and recompression, completing a second stable-state transition.
After two successive transitions, the four-unit stacked structure is ultimately compressed to an overall length equivalent to two unit heights. 
This process clearly demonstrates a cascade-like stable-state transition in the multi-unit system, triggered by local unit instability, and highlights the strongly nonlinear and path-dependent characteristics of the structural response.

\subsubsection{ Dynamics of a compliant crawling robot}

The Kresling origami structure offers the advantage of generating directional actuation through axial stacking of multiple units. 
However, its intrinsic bistability poses challenges for controllability, as the existence of distinct energy wells may lead to abrupt snap-through transitions that are difficult to regulate. 

To address this issue, compliant Kresling designs have been proposed \cite{nayakanti2018twist}.
Following the reported strategy, the modification consists of two key steps. 
First, all mountain creases are removed, resulting in open borders along the corresponding panel edges. 
These previously continuous edges become kinematically unconstrained and are allowed to deform freely, as illustrated in {Figure ~\ref{fig:crawl}} (a). 
Second, the stiffness of the remaining creases is increased to enhance rotational constraint.

We perform simulations on the compliant Kresling mechanism. 
Figure ~\ref{fig:crawl} (b) shows the deformation of the lateral triangular panels when a rotational load is applied to the right annular end face. 
Compared with the rigid counterpart, the compliant structure accommodates geometric incompatibility during folding through bending deformation concentrated near the open borders. 
This bending mechanism replaces the stretching-dominated response observed in the original configuration.
The corresponding energy evolution is presented in Figure ~\ref{fig:crawl} (c). 
The introduction of open borders significantly reduces the energy barrier between stable configurations, primarily due to the substantial decrease in panel stretching energy.

Figure ~\ref{fig:crawl} (d) further illustrates the torque--rotation relationship for increasing crease stiffness. 
When the crease stiffness is small, the structure retains a bistable response characterized by a non-monotonic load--displacement curve. 
As the stiffness increases, the torque--rotation curve gradually becomes monotonic, indicating the disappearance of snap-through behavior and the emergence of a stable stiffness response. 
Such a monotonic response is advantageous for precise actuation and controllable motion in compliant robotic applications.
This demonstrates the advantages of the programmable stability of compliant origami.

\begin{figure}[t]
    \centering
    \includegraphics[
        width=0.9\linewidth,
        trim={0 4cm 0 0},
        clip
    ]{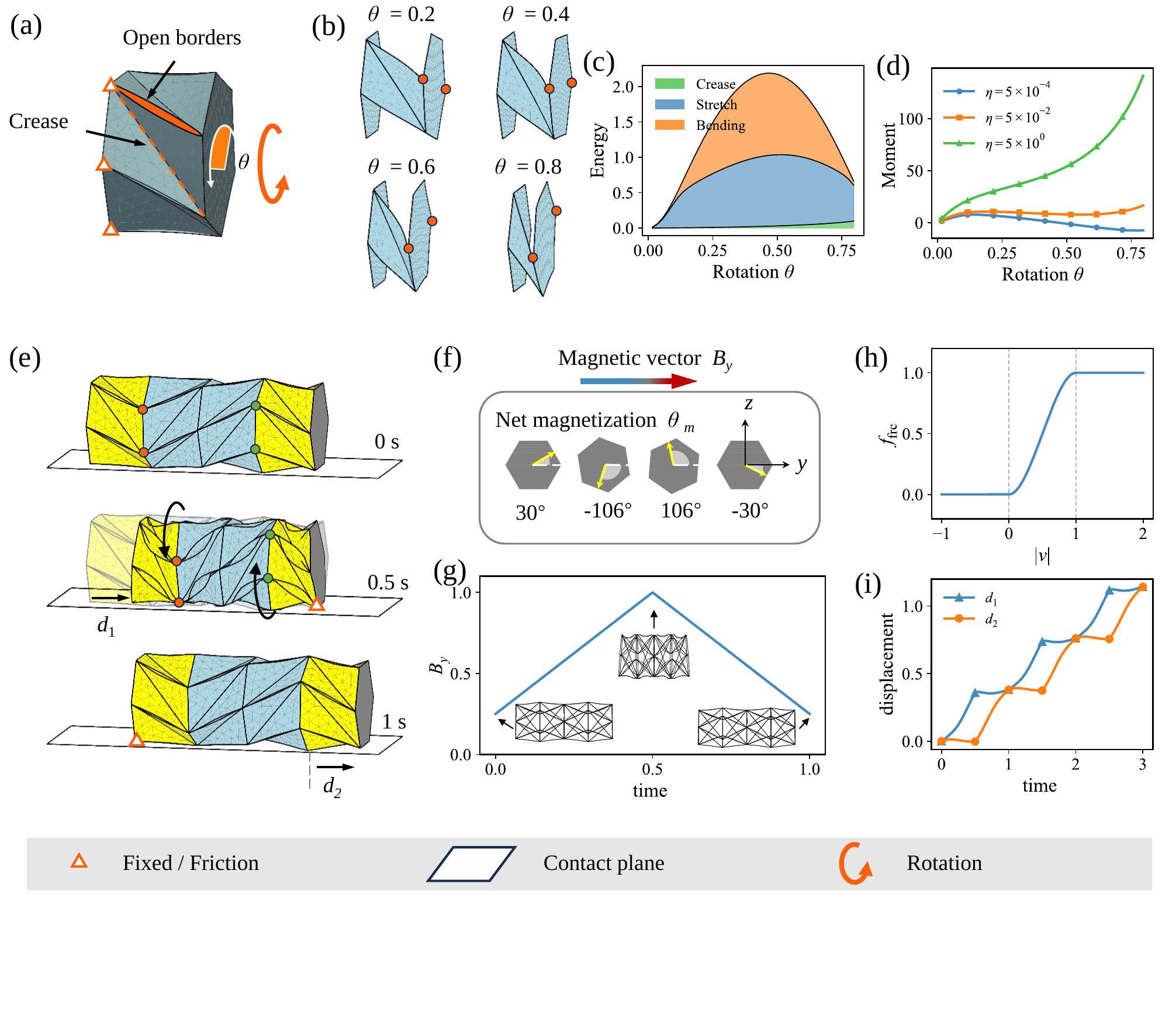}
    \caption{Dynamic locomotion of a compliant crawling robot actuated by Kresling-based units.
(a--d) Mechanical characteristics of a compliant Kresling actuator, including its load--displacement response and multistable behavior.
(e) One complete locomotion cycle of the crawling robot\cite{ze2022soft}.
(f) Schematic of the external magnetic field and orientation of the net magnetization vector in the actuator.
(g) Robot configurations under varying magnetic field strengths.
(h) Definition of the friction coefficient governing foot--ground interaction.
(i) Time histories of the foot displacement during robot locomotion.
}
    \label{fig:crawl}
\end{figure}

Building upon the compliant Kresling design, we simulate the dynamic crawling process of the compliant origami robot proposed in the literature\cite{ze2022soft}. 
As shown in Figure ~\ref{fig:crawl} (e), this ingeniously designed robot consists of four Kresling units stacked in an alternating symmetric and antisymmetric configuration. 
This arrangement enables kinematic decoupling between internal deformation and foot motion. 
Magnetic plates are attached to the transverse hexagonal panels to realize passive torsional actuation under an external magnetic field, which is subsequently converted into axial extension and contraction through the kinematics of the Kresling units. 
Finally, directional crawling is achieved through specially designed feet with anisotropic frictional properties.
In our simulation framework, the elastic and kinetic energies of the compliant origami structure are fully considered, together with magnetic forces, gravitational effects, ground contact, and frictional interactions.

Except for the central panel, four panels of the robot are magnetized. 
The net magnetization directions are illustrated in Figure ~\ref{fig:crawl} (f). 
Each magnetic plate has a magnetization vector inclined by an angle $\theta_m$ relative to the $y$-axis. 
A uniform magnetic field is continuously applied along the $y$ direction. 
The magnetic torque drives the plates to rotate so as to align their magnetization with the external field, generating controlled torsional deformation. 
By adjusting the magnitude of the applied magnetic field, the global configuration of the robot can be regulated. 
As shown in Figure ~\ref{fig:crawl} (g), a weak magnetic field restores the robot to its initial (extended) configuration, whereas a stronger magnetic field induces a folded state.
Forward locomotion is achieved through directional (anisotropic) friction at the robot feet. 
We implement a unidirectional friction model at the contact nodes between the robot feet and the ground. 
As illustrated in Figure ~\ref{fig:crawl} (h), the friction coefficient is defined in a piecewise manner:
\begin{equation}
f_{\text{frc}}(\left| v \right|) =
\begin{cases}
0, & \left| v \right| < 0, \\
3\left| v \right|^2 - 2\left| v \right|^3, & 0 \le \left| v \right| < 1, \\
1, & \left| v \right| \ge 1,
\end{cases}
\end{equation}
where $\left| v \right| = v/\varepsilon_v$ denotes the normalized tangential velocity. 
This formulation ensures $C^1$ continuity across the no-friction, static-friction, and kinetic-friction regimes, thereby avoiding numerical instability. 
The above contact-friction model is applied to the contact nodes on both end faces.

Figure ~\ref{fig:crawl} (i) presents the time histories of the horizontal displacements of two representative foot nodes. 
Each crawling cycle consists of two stages. 
During the first stage ($0$--$0.5\,\mathrm{s}$), the magnetic field strength increases, causing the robot to fold. 
In this phase, leftward motion of the feet is restricted by friction, while rightward motion remains unimpeded. 
As a result, the right foot displacement $d_2$ remains nearly zero, whereas the left foot displacement $d_1$ increases toward the right. 
During the second stage ($0.5$--$1\,\mathrm{s}$), the magnetic field decreases and the robot re-extends. 
The frictional constraint reverses: $d_1$ becomes fixed, and $d_2$ advances to the right. 
Through this alternating deformation and frictional rectification mechanism, the robot achieves net forward crawling over each cycle.

\section{Conclusion}
 \label{sec:conclusion}
 
In this work, we established a unified DDG–based framework for the simulation of robotic origami system spanning the full spectrum from rigid kinematic folding to compliant shell deformation.
Motivated by the need to both understand and engineer compliant origami robots, the proposed formulation bridges traditionally separated modeling paradigms—rigid-folding kinematics, bar-and-hinge abstractions, and continuum shell mechanics—within a single geometrically consistent structure.

By introducing shell and crease elements under a shared mid-edge discretization, and by expressing their mechanics through dual face– and dihedral-angle–based representations, the framework provides a coherent description of bending, folding, and rotation in a unified variational setting.
This consistency enables rigid motion, elastic deformation, and nonlinear stability phenomena to emerge as different regimes of the same underlying formulation, rather than as fundamentally distinct models.
To capture the strongly nonlinear and dynamic behaviors inherent to origami systems, an implicit time integration scheme and environmental interactions—including gravity, contact, friction, and magnetic actuation—were incorporated into the framework.
The resulting model enables the systematic investigation of multi-stability, snap-through transitions, frictional contact, and multiphysics coupling within a robust computational environment.

Through a progressive set of representative examples—including the single-fold (Z-fold), the Miura pattern, the Waterbomb pattern, and the Kresling structure—the framework was validated across increasing levels of geometric complexity and mechanical nonlinearity.
These examples span a spectrum from fundamental folding units to functionally actuated origami robots, covering behaviors ranging from simple geometric reconfiguration to snap-through propulsion and directional locomotion.
For each case, rigid and compliant formulations were examined within the same DDG structure, revealing how mechanical parameter tuning continuously reshapes the system’s response landscape. 
These results demonstrate not only the accuracy and robustness of the approach, but also its capacity to expose the underlying mechanics that govern programmable folding behavior.

Overall, this work provides both a conceptual and computational foundation for compliant dynamic origami systems.
By unifying geometry, elasticity, and dynamics in a single formulation, the framework advances the theoretical understanding of compliant origami mechanics while simultaneously offering a practical design and analysis tool for next-generation programmable origami robots.
Future directions include inverse design strategies, optimization-based pattern synthesis, and integration with model-based control methodologies to further enable mechanically intelligent and adaptive origami robotic systems.

\section*{Supporting Information}

Supplementary information for this article is provided in the attached file.

\section*{Acknowledgements}

B.W. acknowledge the support from the  National Natural Science Foundation of China (No. 12432002, 12402002, 12572073, 62388101), from the Guangdong Basic and Applied Basic Research Foundation (No. 2024A1515010767).
M.L. acknowledges the start-up funding from the University of Birmingham, UK.
W.H. acknowledges the start-up funding from Newcastle University, UK.

\medskip

%
\bibliographystyle{MSP}
\bibliography{sample}
\end{justify}





\end{document}